\documentclass{article}

\usepackage[preprint]{neurips_2026}

\usepackage[utf8]{inputenc} 
\usepackage[T1]{fontenc}    
\usepackage{hyperref}       
\usepackage{url}            
\usepackage{booktabs}       
\usepackage{amsfonts}       
\usepackage{nicefrac}       
\usepackage{microtype}      
\usepackage{xcolor}         
\usepackage{float}
\usepackage{graphicx}
\graphicspath{{figures/}}
\definecolor{llmcolor}{HTML}{FF6B20}
\definecolor{pipelinecolor}{HTML}{5A8A25}
\usepackage{amsmath}
\usepackage{amssymb}
\usepackage{pifont}
\newcommand{\cmark}{\ding{51}}
\newcommand{\xmark}{\ding{55}}
\usepackage{wrapfig}
\renewcommand{\cite}{\citep}
\usepackage{titletoc}
\usepackage{tcolorbox}
\tcbuselibrary{breakable}
\usepackage{array}
\usepackage{enumitem}
\usepackage{listings}
\usepackage{caption}
\title{LLM-Guided ODE Discovery and Parameter Inference from Small-Cohort Aggregate Data}

\author{%
  \parbox{\textwidth}{\centering\normalfont
  \textbf{Hanning Yang}$^{1*}$, \;
  \textbf{Meropi Karakioulaki}$^2$, \;
  \textbf{Lennart Purucker}$^3$, \;
  \textbf{Tim Litwin}$^1$, \;
  \textbf{Cristina Has}$^2$, \;
  \textbf{Moritz Hess}$^1$ \\[0.5em]
  {\footnotesize
  $^1$Institute of Medical Biometry and Statistics, Faculty of Medicine and Medical Center, University of Freiburg, Germany \\
  $^2$Department of Dermatology, Medical Faculty and Medical Center, University of Freiburg, Germany \\
  $^3$Prior Labs, University of Freiburg, Germany \\[0.3em]
  \texttt{hanning.yang@uniklinik-freiburg.de}}
  }
}

\begin{document}

\maketitle

\begin{abstract}
Mechanistic modeling via ordinary differential equations (ODEs) provides interpretable descriptions of complex dynamics and enables inference of underlying mechanisms, which is particularly valuable in clinical settings. However, in rare diseases, both the structure and parameters of the model are typically unknown, while individual-level data is scarce, noisy, heterogeneous, and subject to privacy constraints. In such settings, population-level summary statistics provide a practical privacy-preserving data representation, while capturing heterogeneity further requires modeling parameters as distributions rather than fixed values. Yet no existing method jointly discovers ODE structure and refines parameter distributions solely from summary statistics.
We present AgentODE, an end-to-end framework that addresses this gap. An LLM proposes candidate ODE structures, while a tool-augmented inference agent iteratively refines parameter distributions through a diagnosis--update loop, operating on population-level summary statistics alone.
We evaluate AgentODE on three benchmark problems across different fields and two clinical datasets, including the rare disease recessive dystrophic epidermolysis bullosa (RDEB), with only 231 observations across 46 patients. AgentODE recovers functionally consistent ODE structures across all settings, and experiments on RDEB demonstrates that in sparse and noisy data settings reasoning from summary statistics promotes mechanistically principled structure discovery, whereas baselines with individual-level data access recover implausible structures despite better predictive performance.
AgentODE opens new possibilities for mechanistic modeling of rare diseases directly from population-level summary statistics, where data scarcity and privacy constraints have traditionally limited such analyses\footnote{Code available at \url{https://github.com/HanningYang/AgentODE}.}.
\end{abstract}

\section{Introduction}
\label{sec:intro}
Mechanistic models based on ordinary differential equations (ODEs) provide a principled framework for describing the dynamics of complex systems across domains such as biology, medicine, and engineering\citep{raue2013lessons}. By explicitly encoding mechanistic relationships between variables, ODE models offer interpretability, support reasoning about underlying processes, and enable extrapolation beyond observed data. These properties make ODEs particularly valuable in clinical settings, where understanding disease mechanisms is critical.

However, constructing ODE models typically requires substantial expert knowledge to specify both the functional structure and parameters of the system. In rare diseases, such knowledge is often limited, while available data are typically scarce, noisy, heterogeneous, and subject to privacy constraints~\citep{hilgers2016design}. Capturing this heterogeneity requires modeling parameters as distributions rather than fixed values~\citep{jaqaman2006linking}. These challenges make reliable model identification difficult in practice, even for approaches that rely on fitting individual-level data.
In practice, population-level summary statistics are often used as a pragmatic alternative, as they enable privacy-preserving data sharing and can still provide an informative representation under noise, irregular sampling, and small cohort sizes.

While automatic structure discovery methods such as symbolic regression~\citep{makke2024interpretable} and sparse identification~\citep{brunton2016discovering} can recover governing equations from data, they rely on dense individual-level data for model fitting and evaluation. Recent LLM-based ODE discovery approaches~\citep{holt2024data, du2024llm4ed, yang2026think, bideh2026llm} extend this paradigm by leveraging scientific knowledge embedded in LLMs to propose candidate structures, but similarly evaluate fitness against individual-level data.
Meanwhile, population-level approaches such as nonlinear mixed-effects models~\citep{lavielle2014mixed} address data sparsity by pooling information across individuals, but require a pre-specified model structure.
Together, these limitations expose a critical gap: no 
existing method jointly discovers ODE structure and refines 
parameter distributions from population-level summary statistics 
alone.

Large language models have demonstrated strong capabilities in 
scientific knowledge retrieval, mathematical reasoning, and code 
generation~\citep{romera2024mathematical, shojaee2024llm}, as well 
as visual interpretation of complex data~\citep{achiam2023gpt} and 
proposal of informative prior distributions in scientific 
contexts~\citep{riegler2025using}. 
These capabilities map naturally onto the core challenges of ODE discovery in data-scarce settings. Scientific knowledge retrieval and code generation enable the proposal of mechanistically grounded ODE structures without explicit expert input. Visual reasoning allows the model to interpret patterns from summary statistics presented as visual summaries. The ability to propose informative prior distributions motivates modeling parameters as distributions to capture population-level heterogeneity. Finally, the agentic paradigm enables LLMs to interleave reasoning and tool use, iteratively diagnosing and refining outputs based on feedback~\citep{yao2022react}. 

\begin{figure}[htbp]
  \centering
  \includegraphics[width=\linewidth]{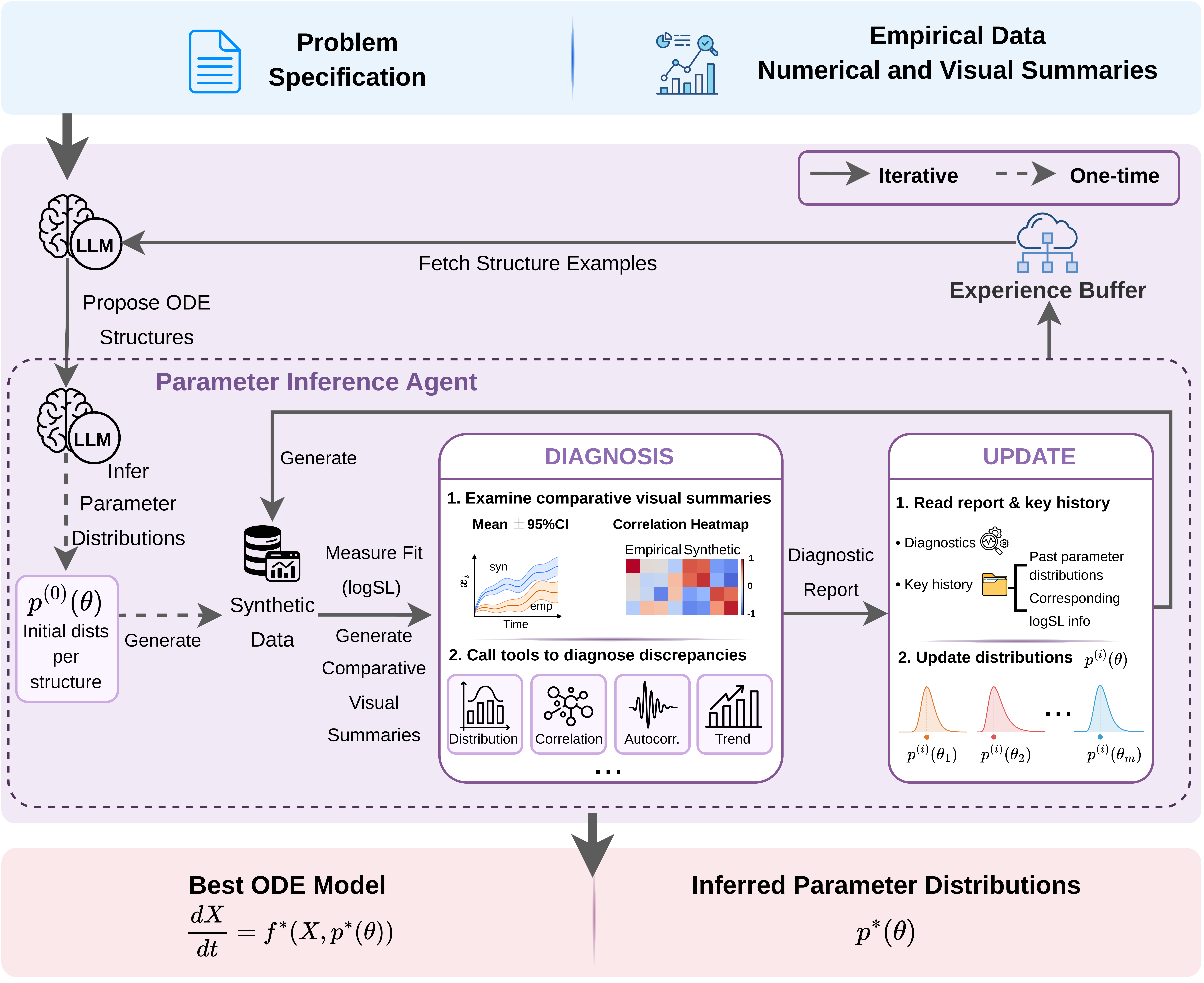}
  \caption{Overview of \textbf{AgentODE}. Inputs consist of a problem specification and empirical summaries. An LLM proposes candidate ODE structures. For each structure, the agent performs an \textbf{initial inference} to obtain starting parameter distributions, from which synthetic data are generated, evaluated using logSL, and used to produce comparative visual summaries against empirical data. The agent then refines the parameters through a \textbf{diagnosis--update} loop. In \textbf{diagnosis}, discrepancies are identified from comparison visuals and quantified using statistical tools to produce a diagnostic report. In \textbf{update}, the agent reads this report and learns from past parameter distributions and corresponding logSL information to update the parameter distributions. An \textbf{experience buffer} stores structures and parameter distributions to guide future structure proposals. AgentODE outputs the best ODE model $f^*$ and inferred parameter distributions $p^*(\theta)$.}
  \label{fig:overview}
\end{figure}

Building on these insights, we propose AgentODE (Figure~\ref{fig:overview}), an end-to-end framework that jointly discovers ODE structures and refines parameter distributions from population-level summary statistics. 
At its core, AgentODE consists of two coupled feedback loops mediated by the LLM: an outer structure-discovery loop that proposes and scores candidate ODE structures based on discrepancies between synthetic and empirical summary statistics, and an inner parameter-inference loop that iteratively refines parameter distributions by diagnosing and reducing these discrepancies.
The process begins with the LLM proposing candidate ODE structures. For each structure, the parameter inference agent refines parameter distributions through an iterative diagnosis--update loop. At each iteration, the current distributions are used to simulate synthetic data, from which summary statistics are computed and compared with empirical summary statistics to identify discrepancies. The agent then reasons over these discrepancies together with memory of previous high-performing iterations to update the parameter distributions.
An experience buffer with dynamic memory stores discovered results and provides examples to guide subsequent ODE structure proposals.

We evaluate AgentODE on three synthetic benchmark problems spanning materials science, pharmacokinetics, and molecular biology, as well as two clinical datasets: acute kidney injury (AKI) and recessive dystrophic epidermolysis bullosa (RDEB), a rare skin disorder with only 231 observations across 46 patients. Using GPT-5.2 as the backbone LLM, AgentODE achieves performance comparable to baselines with full trajectory access, while operating on population-level summary statistics. On RDEB, we show that reasoning from population-level summary statistics provides a more robust signal for structure discovery: by condensing population-level patterns while mitigating individual-level noise, summary statistics guide the LLM toward the true underlying mechanism, whereas baselines with access to individual-level data, despite achieving lower prediction error, discover mechanistically implausible or degenerate structures. This demonstrates that in sparse and noisy clinical settings, predictive performance alone is an insufficient criterion for evaluating mechanistic model quality.
\textbf{Our contributions} are as follows:
\begin{itemize}[leftmargin=*]
    \item We propose AgentODE, the first framework that jointly discovers ODE 
    structures and refines parameter distributions from population-level 
    summary statistics.

    \item We demonstrate that, in sparse and noisy clinical settings, summary statistics can serve as a robust and informative data representation to guide mechanistic model inference, enabling recovery of underlying dynamical behavior while mitigating overfitting to individual-level noise.

    \item We show that the jointly inferred structure and parameter 
    distributions can be used to generate synthetic patient populations 
    that capture observed heterogeneity.
\end{itemize}

\section{Related Work}
\label{sec:related}

\textbf{Agentic reasoning with LLMs.}\quad 
The ReAct framework~\citep{yao2022react} established the paradigm of 
interleaving reasoning and tool use in LLM agents. Reflexion~\citep{shinn2023reflexion} extended 
this idea by incorporating reflection over past failures stored in an 
episodic memory buffer, allowing agents to improve decisions across 
trials. Related approaches have further explored 
structured reasoning and planning in LLMs~\citep{yao2023tree, wang2023plan}. 
LLMs have also demonstrated strong capabilities in scientific tasks such as knowledge 
retrieval~\citep{romera2024mathematical, shojaee2024llm}, visual interpretation of complex data~\citep{achiam2023gpt}, and the proposal 
of informative prior distributions~\citep{riegler2025using}. Building on these advances, AgentODE applies agentic reasoning to mechanistic model discovery, where iterative feedback from summary statistics guides both structure discovery and parameter inference.

\textbf{Equation and ODE structure discovery.}\quad Classical 
methods including symbolic regression~\citep{makke2024interpretable} 
and sparse identification~\citep{brunton2016discovering} recover 
governing equations from data but require individual-level 
data and rely on predefined operator sets or function libraries. 
Related advances in scientific machine learning, such as 
neural ODEs~\citep{chen2018neural} and universal differential 
equations~\citep{rackauckas2020universal}, enable flexible modeling 
of dynamical systems but rely on uninterpretable neural components rather than explicit mechanistic structures.
LLM-based approaches such as LLM-SR~\citep{shojaee2024llm} and 
LaSR~\citep{grayeli2024symbolic} partially address this gap by leveraging 
scientific knowledge embedded in LLMs to guide equation discovery through 
search or reusable concept libraries. Recent approaches 
including LLM4ED~\citep{du2024llm4ed}, D3~\citep{holt2024data}, 
KeplerAgent~\citep{yang2026think} leverage LLMs to propose ODE structures but still depend on individual-level data for evaluation. 
Population-level approaches such as nonlinear mixed-effects 
models~\citep{lavielle2014mixed} pool information across individuals but require a predefined ODE structure. 
To our knowledge, AgentODE is the first framework to jointly discover ODE structure and 
refine parameter distributions from population-level summary statistics.

\textbf{Simulation-based inference.}\quad 
Simulation-based inference encompasses likelihood-free parameter estimation methods~\citep{cranmer2020frontier}. Classical approaches such as approximate Bayesian computation~\citep{beaumont2019approximate} compare empirical and synthetic data through summary statistics, while more recent neural methods such as sequential neural likelihood~\citep{papamakarios2019sequential} and neural posterior estimation~\citep{greenberg2019automatic} learn posterior distributions using neural density estimators. These approaches can require large simulation budgets and can produce less interpretable outputs.
The synthetic likelihood~\citep{wood2010statistical} defines a likelihood over summary statistics, enabling parameter inference by comparing synthetic and empirical summary statistics.
This aligns with our setting, where parameter inference relies on summary statistics.

Table~\ref{tab:method_comparison} compares existing methods and AgentODE across key capabilities. LLM-SR$^\dagger$ denotes our adaptation for ODE discovery. AgentODE is the only method that jointly discovers ODE structure, infers parameter distributions $p(\theta)$, and operates on population-level summary statistics.

\begin{center}
\small
\captionof{table}{Method comparison across key capabilities. $^\dagger$Adaptations for ODE discovery.}
\label{tab:method_comparison}
\vspace{0.5em}
\begin{tabular}{lcccc}
\toprule
\textbf{Method} & 
\textbf{Discovers structure} & 
\textbf{Infers $p(\theta)$} & 
\textbf{Population-level summaries} & 
\textbf{Interpretable} \\
\midrule
SINDy & \cmark & \xmark & \xmark & \cmark \\
PySR & \cmark & \xmark & \xmark & \cmark \\
LLM-SR$^\dagger$ & \cmark & \xmark & \xmark & \cmark \\
Neural ODE & \xmark & \xmark & \xmark & \xmark \\
\midrule
\textbf{AgentODE} & \cmark & \cmark & \cmark & \cmark \\
\bottomrule
\end{tabular}
\end{center}
\section{AgentODE}
\label{sec:method}

AgentODE is designed for clinical settings such as rare diseases, operating on population-level summary statistics rather than directly using individual-level data. The framework consists of two tightly coupled components: an LLM-guided ODE structure discovery module (the outer loop; Section~\ref{sec:structure}) and a tool-augmented parameter inference agent (the inner loop; Section~\ref{sec:agent}). 
The structure discovery module proposes candidate ODE structures, which are evaluated by the parameter inference agent through iterative refinement of parameter distributions. This evaluation is based on discrepancies between synthetic and empirical summary statistics, which guide subsequent structure proposals, forming a closed-loop process that jointly improves both structure discovery and parameter inference. We first formally introduce the problem setting.

\subsection{Problem Formulation}
\label{sec:problem}

Our goal is to infer an ODE structure $f^*$ and parameter distribution 
$p^*(\theta)$ such that synthetic data generated by the model produce 
summary statistics consistent with the empirical data, thereby capturing 
the underlying dynamical patterns of the system.

We consider ODEs of the form $\frac{dX}{dt} = f(X, \theta)$, where $X \in \mathbb{R}^d$ denotes the state variables and $\theta \in \mathbb{R}^m$ the model parameters. 
To capture inter-individual heterogeneity, we assume independent LogNormal distributions for parameters, $\theta_j \sim \mathrm{LogNormal}(\mu_j, \sigma_j^2)$.
Given empirical data $\mathcal{X} = \{x_i(t)\}_{i=1}^{N}$ consisting of 
$N$ individual trajectories, we represent the data using population-level 
summary statistics $s_{\mathrm{emp}} = \phi(\mathcal{X}) \in \mathbb{R}^q$,
where $\phi$ denotes a summary-statistic mapping and $q$ is the number 
of summary statistics.
For a candidate structure $f$ and parameter distribution $p(\theta)$, 
synthetic data $\tilde{\mathcal{X}}(f, p(\theta))$ are generated, 
from which summary statistics $s_{\mathrm{syn}} = \phi(\tilde{\mathcal{X}}(f, p(\theta)))$ are computed.
The objective is to infer $f^*$ and $p^*(\theta)$ by minimizing a discrepancy 
between empirical and synthetic summary statistics, $(f^*, p^*(\theta)) = \arg\min_{f,\,p(\theta)} 
\mathcal{D}\big(s_{\mathrm{syn}}, s_{\mathrm{emp}}\big)$,
where $\mathcal{D}(\cdot,\cdot)$ denotes a discrepancy function defined 
over summary statistics, as detailed in Sections~\ref{sec:stats} 
and~\ref{sec:scoring}.

\subsection{Summary Statistics}
\label{sec:stats}

Summary statistics support both structure discovery and parameter inference. For structure discovery, we use a broad set of statistics capturing distributions, autocorrelation, stationarity, entropy, scaling, trend, volatility, and cross-variable dependencies~\citep{fulcher2018feature, fulcher2014highly, bandt2002permutation, kantelhardt2001detecting, hyndman2018forecasting}, enabling informative comparison between candidate models.
For parameter inference, we use a focused subset of the summary statistics used for structure comparison to provide stable signals for likelihood estimation. To better capture the dynamics of sparse and noisy clinical data, we prioritize distributional shape, temporal autocorrelation, population-level trends, and cross-variable dependencies, which together reflect the most informative aspects of population dynamics while ensuring consistency between parameter inference and structure comparison (complete set of statistics is provided in Appendix~\ref{sec:appendix_stats}).

\subsection{Scoring}
\label{sec:scoring}

AgentODE employs two complementary metrics: log synthetic likelihood (logSL) for parameter inference within an ODE structure, and mean normalized summary discrepancy (MNSD) for comparison across ODE structures.
LogSL evaluates parameter fit within a structure but is not comparable across structures, as each structure has its own parameterization and induces a different distribution over summary statistics. We therefore introduce MNSD for consistent comparison across structures.

\textbf{Scoring within a structure (logSL).}\quad 
Given a candidate structure $f$ and parameter distribution $p(\theta)$ parameterized by $\psi = \{(\mu_j, \sigma_j)\}_{j=1}^m$, we perform $N_{\mathrm{sim}}$ simulations by drawing independent parameter vectors $\{\theta^{(n)}\}_{n=1}^{N_{\mathrm{sim}}}$ from $p(\theta)$. Each simulation generates a population of synthetic trajectories $\tilde{\mathcal{X}}^{(n)}(f, \theta^{(n)})$, from which summary statistics $s_{\mathrm{syn}}^{\mathrm{SL},(n)} = \phi_{\mathrm{SL}}(\tilde{\mathcal{X}}^{(n)}(f, \theta^{(n)}))$ are computed.
Following \citep{wood2010statistical}, we evaluate model fit using the log synthetic likelihood $\log \mathrm{SL}(\psi) = -\tfrac{1}{2} \left[
\log |\hat{\Sigma}_\psi| + 
(s_{\mathrm{emp}}^{\mathrm{SL}} - \hat{\mu}_\psi)^\top 
\hat{\Sigma}_\psi^{-1} 
(s_{\mathrm{emp}}^{\mathrm{SL}} - \hat{\mu}_\psi)
\right]$,
where $\hat{\mu}_\psi$ and $\hat{\Sigma}_\psi$ denote the mean and covariance of $\{s_{\mathrm{syn}}^{\mathrm{SL},(n)}\}_{n=1}^{N_{\mathrm{sim}}}$.
The logSL score is maximized during parameter inference.

\textbf{Scoring across structures (MNSD).}\quad 
To compare candidate ODE structures, we quantify discrepancies between empirical and synthetic summary statistics using a normalized absolute difference, which we term the mean normalized summary discrepancy (MNSD): 
$
\mathrm{MNSD} = \frac{1}{|\mathcal{I}|}\sum_{i \in \mathcal{I}}\frac{\left|s^{\mathrm{MNSD}}_{\mathrm{emp},i}-s^{\mathrm{MNSD}}_{\mathrm{syn},i}\right|}{\mathrm{IQR}^{\mathrm{emp}}_i}.
$

Here, \(s^{\mathrm{MNSD}}_{\mathrm{emp},i}\) denotes the \(i\)-th empirical summary statistic, and \(s^{\mathrm{MNSD}}_{\mathrm{syn},i}\) is computed from synthetic data generated using the inferred parameter distribution for the candidate structure. $\mathrm{IQR}^{\mathrm{emp}}_i$ is the empirical interquartile range of that statistic~\citep{law1986robust}. The index set $\mathcal{I}$ indexes the summary statistics used for structure comparison. Lower MNSD indicates better agreement between empirical and synthetic summary statistics.

\subsection{ODE Structure Discovery}
\label{sec:structure}

\textbf{Base framework: LLM-SR.}\quad 
We build on LLM-SR~\citep{shojaee2024llm}, a symbolic regression framework that iteratively generates, evaluates, and refines candidate symbolic equations using LLMs. At each iteration, the LLM proposes equation hypotheses conditioned on the problem description, which are then evaluated based on their fit to individual-level data. The candidate equations and their corresponding fit scores are stored in an experience buffer, which serves as a dynamic memory to guide subsequent proposals. The buffer adopts an island-based model~\citep{tanese1989distributed} to maintain diversity and avoid local optima.

\textbf{Extension to ODE discovery.}\quad 
We extend this paradigm to dynamical systems by replacing symbolic regression with ODE structure discovery. 
Unlike LLM-SR, which directly evaluates candidate equations using individual-level data, our framework evaluates ODE structures via a parameter inference agent~\ref{sec:agent}. For each candidate structure, the agent refines parameter distributions using logSL, after which the structure is scored using MNSD.

\subsection{Parameter Inference Agent}
\label{sec:agent}

Given a candidate ODE structure $f$, the agent operates in two phases: an initial inference producing a starting parameter distribution, followed by an iterative diagnosis--update loop that refines the parameter distribution $p(\theta)$, guided by logSL measuring agreement between $s_{\mathrm{syn}}^{\mathrm{SL}}$ and $s_{\mathrm{emp}}^{\mathrm{SL}}$.

\textbf{Visual summaries.}\quad
We construct visual summaries from population-level data in a structured and interpretable form to represent temporal dynamics, heterogeneity, and inter-variable relationships. These include (i) mean trajectories with 95\% confidence intervals over time, (ii) mean trajectories with 95\% confidence intervals stratified by initial-condition quartiles, and (iii) heatmaps of Spearman correlations of first differences between variables.
These summaries are used as empirical inputs for initial parameter inference. During diagnosis, empirical and synthetic versions of these summaries are compared to identify discrepancies that guide parameter updates.

\textbf{LLM Initial Inference}\quad
Given the problem specification, candidate ODE structure $f$, and empirical visual summaries, the LLM proposes an initial parameter distribution $p^{(0)}(\theta)$. Details in Appendix~\ref{sec:appendix_prompts_init}.

\textbf{Diagnosis.}\quad
The agent diagnoses the current best-performing parameter 
distribution, as measured by logSL. It first reasons from 
the problem context and the ODE structure $f$ to infer 
expected system dynamics, including trajectory behavior, 
heterogeneity, and relationships between variables. 
These expectations are grounded by comparing empirical and synthetic 
visual summaries to identify discrepancies in the simulated 
dynamics. 
Beyond the most apparent mismatches, the agent reasons about their underlying causes in the context of $f$ and the problem domain, linking them to parameter 
roles and the dynamical assumptions implied by the current 
distribution. The same summary statistics set used for structure 
scoring are provided as tools to the agent, which selectively 
evaluates them to quantify discrepancies and support its 
interpretation. The results are compiled into a structured 
diagnostic report that characterizes each failure mode by its 
severity, affected variables, and supporting statistical evidence.
Details in Appendix~\ref{sec:appendix_prompts_diagnosis_update}.

\textbf{Update.}\quad
Given the diagnostic report and past parameter distributions with high logSL scores, the agent updates the parameter distribution.
First, it examines the diagnosed failure modes and reasons in the context of the ODE structure $f$ and the problem domain to identify which parameters govern the affected dynamics and what directional changes may be required.
It then reviews past parameter settings and their associated logSL outcomes to understand how previous configurations influenced model fit, informing the next update and avoiding repeating ineffective configurations.
\section{Experiments and Results}
\label{sec:experiments}

\subsection{Benchmarks and Clinical Datasets}
\label{sec:datasets}
We evaluate AgentODE on three synthetic benchmarks and two clinical datasets, focusing on ODE structure recovery and inferred parameter distributions.


\textbf{Benchmarks.}\quad 
We evaluate AgentODE on three benchmark 
problems spanning materials science, pharmacokinetics, and molecular 
biology. To ensure evaluation 
beyond memorization, we assess recitation risk through contamination 
analysis (Table~\ref{tab:contamination} in Appendix~\ref{sec:appendix_benchmarks}).
The \textbf{Apoptosis} system models programmed cell death 
dynamics across three coupled molecular species in molecular 
biology~\citep{bideh2026llm}. The \textbf{Polymer DA Cross-linking} 
system models Diels--Alder reaction kinetics across four coupled 
chemical species in materials science~\citep{bideh2026llm}. The 
\textbf{PKPD-Immune} system couples two-compartment 
pharmacokinetics with immune and effector cell dynamics across 
four state variables.
AgentODE is designed to operate over heterogeneous populations, 
requiring that benchmark systems accommodate individual-level 
variability. We therefore sample initial conditions from 
domain-grounded distributions for all systems. For parameters, 
we introduce inter-individual variability where meaningful: 
the \textbf{PKPD-Immune} system naturally accommodates 
inter-individual variability in pharmacokinetics and immune 
response, and we assign log-normal parameter distributions 
accordingly. For the remaining systems, parameters are fixed 
following prior work~\citep{bideh2026llm}. 
Details in Appendix~\ref{sec:appendix_benchmarks}.

\textbf{Clinical Datasets.}\quad
We evaluate AgentODE on two clinical datasets. The first is an 
acute kidney injury (AKI) cohort extracted from MIMIC-IV 
3.1~\citep{johnson2023mimic}, comprising 353 admissions (trajectories) from 
patients aged 50--64 years, with irregularly spaced measurements 
of creatinine, blood urea nitrogen (BUN), and potassium over 
the first 168 hours after admission. 
Measurements are irregularly spaced with variable numbers of observations per trajectory.
The second dataset comprises 46 patients with severe recessive 
dystrophic epidermolysis bullosa 
(RDEB)~\citep{bardhan2020epidermolysis, reimer2020natural, 
karakioulaki2026systemic}, a rare genetic skin disease, with 
231 total observations across up to 144 months since first 
hospital visit, a mean of 5 observations per patient, and 16 
patients with only 2 time points. We use measurements of serum 
albumin, log-transformed C-reactive protein 
(log$_{10}$(CRP$+$1)), hemoglobin, and BMI relative to 
age- and sex-matched WHO reference (BMI\textsubscript{rel}).

\subsection{Experimental Setup}
\label{sec:setup}
We compare AgentODE against state-of-the-art structure discovery baselines spanning sparse regression (\textbf{SINDy}~\citep{brunton2016discovering}), evolutionary symbolic regression (\textbf{PySR}~\citep{cranmer2023interpretable}), and LLM-guided symbolic regression (\textbf{LLM-SR}~\citep{shojaee2024llm}), as well as \textbf{Neural ODE}~\citep{chen2018neural} as a black-box baseline. For PySR, we fit one independent symbolic regression model per state variable and assemble them into a system of equations. LLM-SR$^\dagger$ denotes our adaptation of LLM-SR to ODE 
discovery, where the LLM proposes ODE structure candidates governing 
all state variables jointly, and parameters are subsequently 
fitted independently for each trajectory using BFGS, providing 
a reference with full access to individual-level data.
SINDy and PySR are not evaluated on clinical datasets, as they require preprocessing steps such as interpolation and derivative estimation that introduce assumptions incompatible with our setting.
We additionally include two summary-level ablations: \textbf{LLM-based parameter initialization ($k=5$)}, which queries the LLM 5 times given the ODE structure and problem context to propose parameter distributions and selects the one with the highest logSL for evaluation, and \textbf{AgentODE w/o iterative refinement}, which removes the diagnosis--update loop from AgentODE.
For parameter inference, AgentODE uses a bounded iteration budget with early stopping (10 iterations per structure with patience 5).
LLM-SR$^\dagger$ is evaluated with over 1000 sampled candidates. AgentODE and its ablations use approximately 500 samples per experiment.
All other baselines follow their original setups or are tuned to achieve their best performance.
Implementation details in Appendix~\ref{sec:appendix_implementation}.

\subsection{Quantitative Results}
\label{sec:results_quantitative}

\textbf{Performance}\quad
Table~\ref{tab:results} reports RMSE across all datasets. Among summary-level methods, AgentODE achieves the best performance on four of five datasets. On benchmarks, AgentODE performs competitively with full-trajectory methods, particularly on PKPD, where it achieves an RMSE of 0.064 compared to 0.685 for Neural ODE. On clinical datasets, AgentODE substantially outperforms Neural ODE on AKI (2.56 vs 6.162) and achieves comparable performance to LLM-SR$^\dagger$.
However, in sparse and noisy clinical settings, predictive error alone is insufficient for evaluating mechanistic models: lower RMSE may reflect overfitting to noisy trajectories rather than recovery of clinically meaningful dynamics. Across datasets, AgentODE consistently recovers more compact and mechanistically meaningful ODE structures, despite operating solely on summary-level statistics. Qualitative comparisons of the discovered systems are provided in Appendix~\ref{sec:appendix_results}.

\begin{table}[t]
\centering
\caption{
\textbf{RMSE}: averaged across state variables (lower is better).
\textbf{Bold} indicates best performance within each group.
\textbf{N/A}: not applicable due to data requirements.
\textbf{Invalid}: proposed parameter distributions yielded unstable (e.g., NaN/Inf) or physically unrealistic synthetic trajectories.
For RDEB, results are reported in-sample due to limited cohort size.
LLM backbone: \texttt{GPT-5.2}.}
\label{tab:results}
\small
\setlength{\tabcolsep}{4pt}
\begin{tabular}{lccc|cc}
\toprule
& \multicolumn{3}{c}{\textbf{Benchmarks}} 
& \multicolumn{2}{c}{\textbf{Clinical}} \\
\cmidrule(lr){2-4} \cmidrule(lr){5-6}
\textbf{Method} &
\textbf{Apoptosis} &
\textbf{Polymer} &
\textbf{PKPD} &
\textbf{AKI} &
\textbf{RDEB} \\
\midrule
\multicolumn{6}{l}{\textit{Full trajectory access}} \\
SINDy & \textbf{0.004} & 0.543 & 1.251 & N/A & N/A \\
PySR & 0.045 & \textbf{0.196} & 0.981 & N/A & N/A \\
LLM-SR$^\dagger$ & 0.024 & 0.508 & \textbf{0.015} & \textbf{2.583} & 0.532 \\
Neural ODE & 0.134 & 0.257 & 0.685 & 6.162 & \textbf{0.528} \\
\midrule
\multicolumn{6}{l}{\textit{Summary-level only}} \\
LLM-based parameter initialization ($k=5$) & Invalid &Invalid & 0.53 & Invalid& Invalid\\
AgentODE w/o iterative refinement & 0.111 & 1.81 & 0.259 & 3.803 & \textbf{0.628} \\
\textbf{AgentODE} & \textbf{0.079} & \textbf{0.689} & \textbf{0.064} & \textbf{2.56} & 0.644 \\
\bottomrule
\end{tabular}
\end{table}

\begin{figure}[t]
    \centering
    \includegraphics[width=\textwidth]{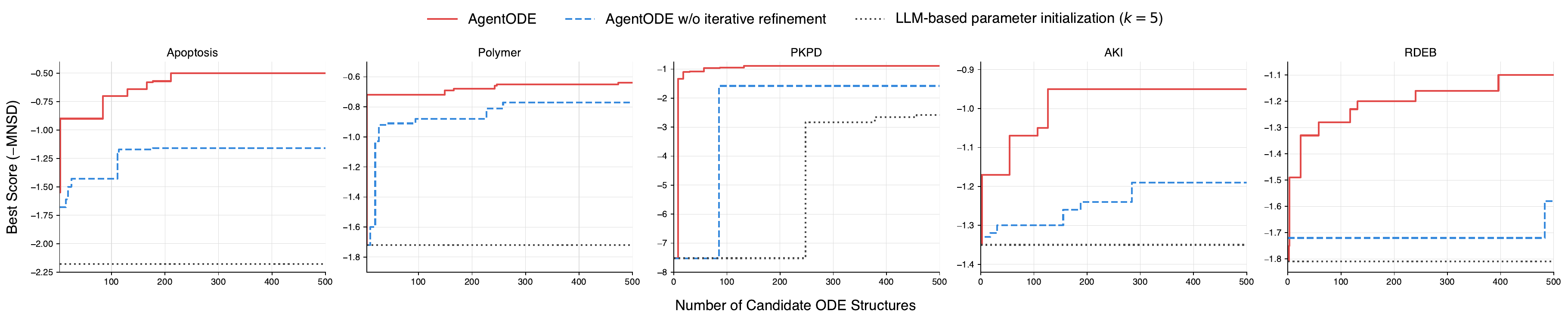}
    \caption{Ablation study comparing AgentODE, AgentODE w/o iterative refinement, and LLM-based parameter initialization ($k=5$) across five datasets. Each panel shows the best score ($-$MNSD, higher is better) over the number of candidate ODE structures evaluated. AgentODE consistently achieves higher scores and improves more efficiently than both ablations across all datasets.}
    \label{fig:ablation}
\end{figure}

\textbf{Efficiency}\quad
Figure~\ref{fig:ablation} shows the best score trajectory over structure discovery iterations for AgentODE, AgentODE w/o iterative refinement, and LLM-based parameter initialization ($k=5$) across all datasets. AgentODE exhibits more consistent improvement over the search, with more frequent score improvements and fewer extended plateaus compared to both ablations, indicating that iterative refinement plays a critical role in sustaining productive exploration. The gap between methods is most pronounced on clinical datasets, where AgentODE achieves substantially higher final scores within the same iteration budget.

\textbf{Analysis.}\quad 
The ablation results reveal two complementary roles of the framework design. 
First, visual summaries provide an effective inductive signal for inferring valid parameter distributions, as evidenced by the Invalid results (4/5) for LLM-based parameter initialization (k=5) across datasets.
This effect is most pronounced for Apoptosis and Polymer, where parameters are fixed and must be inferred from data patterns, and in clinical settings where visual summaries provide a direct representation of temporal patterns, without which the model produces invalid outcomes on AKI and RDEB.
For PKPD, where parameters follow distributions, embedded domain knowledge enables reasonable initialization, though visual summaries still lead to substantial improvement (0.064 vs 0.259). 
Second, diagnosis--update loop consistently improves final performance compared to AgentODE w/o iterative refinement, achieving substantially lower RMSE on Polymer (0.689 vs 1.81), PKPD (0.064 vs 0.259), and AKI (2.56 vs 3.803). 
Figure~\ref{fig:ablation} further shows that refinement leads to more sustained improvement during structure discovery.
Finally, benchmark problems comprise clean, fully observed synthetic trajectories, where full trajectory access provides a natural advantage, whereas clinical datasets are scarce, noisy, and heterogeneous, precisely where summary-level inference is particularly valuable. The RDEB case study illustrates this concretely: AgentODE w/o iterative refinement and LLM-SR$^\dagger$ achieve lower RMSE yet recover mechanistically implausible or degenerate ODE structures (causal comparison in Figure~\ref{fig:rdeb_causal}), suggesting that methods fitting directly to noisy individual-level data overfit to noise rather than recovering the true underlying mechanism. Since high-scoring structures guide future proposals through the experience buffer, such degenerate structures can mislead the entire discovery process, as evidenced by the final structures discovered by these baselines ( Appendix~\ref{sec:appendix_ode_rdeb}). Reasoning from population-level summary statistics provides a more robust signal that mitigates this risk.

\subsection{Qualitative Results}
\label{sec:results_qualitative}

\begin{figure}[t]
    \centering
    \includegraphics[width=\textwidth]{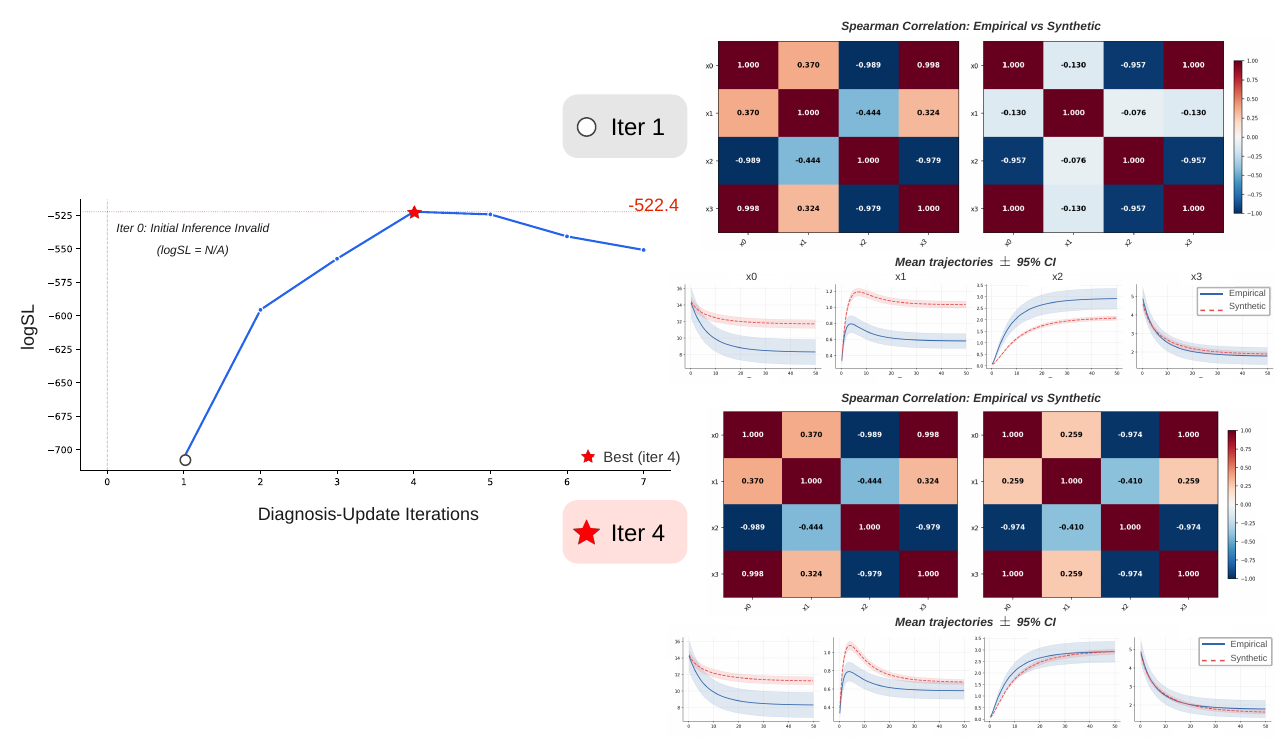}
    \caption{\textbf{Diagnosis-update iterations} for an ODE structure on polymer. \textbf{Left:} logSL score across iterations, with the best iteration marked. \textbf{Right:} Spearman correlation heatmaps and mean trajectories $\pm$ 95\% CI comparing empirical and synthetic data at iteration 1 and the best iteration.}
    \label{fig:logsl_curve}
\end{figure}

\textbf{Diagnosis--Update Loop}\quad
Figure~\ref{fig:logsl_curve} illustrates the diagnosis--update loop for a candidate ODE structure on the Polymer benchmark. The logSL score improves substantially from iteration 1 to the best iteration 4. This improvement is reflected in the visual summaries: at iteration 1, the synthetic correlation structure poorly matches the empirical heatmap, and the mean trajectories with 95\% CI exhibit clear misalignment. By iteration 4, the synthetic correlation structure closely matches the empirical one, and the mean trajectories with 95\% CI align better with the empirical data.

\textbf{Analysis.}\quad 
The diagnosis--update loop improves parameter distributions, as reflected by consistent logSL improvements in early iterations (Appendix~\ref{sec:appendix_prompts_diagnosis_update}). However, the non-monotonic behavior, where the score slightly decreases after the best iteration, highlights an inherent limitation: parameter distributions are updated through LLM-based reasoning over few iterations, rather than gradient-based optimization, and thus monotonic convergence is not guaranteed. 
This is further compounded by parameter non-identifiability~\citep{raue2009structural}, where multiple parameter configurations may yield similarly plausible summary statistics. 
Nevertheless, for a substantial subset of parameters, the inferred distributions fall within plausible ranges and are more concentrated than empirically fitted distributions obtained via trajectory-level optimization (Figure~\ref{fig:param_dist_rdeb}). This behavior is consistent with LLM-based inference concentrating probability mass around plausible values when reasoning from population-level summary statistics, as further supported by correlation heatmaps (Figure~\ref{fig:diff_corr_rdeb}).

\section{Conclusion}
\label{sec:conclusion}

We introduced AgentODE, an end-to-end framework that jointly discovers ODE structures and infers parameter distributions from population-level summary statistics. 
Across benchmarks and clinical datasets, AgentODE achieves performance comparable to baselines that rely on individual-level data. 
The rare disease RDEB case study demonstrates that reasoning from summary statistics provides a robust signal for structure discovery, mitigating overfitting to noisy clinical observations. AgentODE recovers a mechanistically consistent structure aligned with an independently developed expert model, whereas baselines with full data access produce structurally less coherent models that deviate from clinically meaningful relationships. 
Crucially, operating on population-level summary statistics provides a principled privacy-preserving way to mechanistic modeling for rare disease research. 

\textbf{Limitations and Broader Impact.}\quad 
AgentODE's performance relies on knowledge embedded in the LLM and decreases with increasing ODE complexity. It is not intended to replace numerical optimization with full trajectories; parameter identifiability and distribution refinement under data scarcity remain fundamental challenges. Rather, AgentODE provides a practical approach for settings where data are limited, noisy, heterogeneous, or privacy-constrained.
\\
AgentODE has direct applications in biomedical research for under-studied diseases where privacy constraints limit access to individual-level data. Recovered structures and distributions should be interpreted as functionally consistent with observed population behavior rather than ground truth, and validated by domain experts before clinical use.

\textbf{Future Work and Outlook.}\quad 
AgentODE opens several promising directions. On the methodological side, Meta-Harness--style approaches~\citep{lee2026meta} could enable automatic optimization of the agentic pipeline, improving parameter distribution refinement without manual design. Extending the framework to a society of thought~\citep{evans2026agentic}, where multiple specialized agents jointly reason over structure and parameters, could further strengthen the diagnosis--update loop. 
\\
We aim to bring AgentODE into practice to help clinicians understand and cure rare diseases in settings with data scarcity and privacy constraints. 




\begin{ack}
Funded by the Deutsche Forschungsgemeinschaft (DFG, German Research Foundation) – Project-ID 499552394 – SFB 1597.
\end{ack}

\bibliographystyle{plainnat}
\bibliography{references}
\newpage

\appendix
\section*{Appendix}
\startcontents[appendix]
\printcontents[appendix]{l}{1}{\setcounter{tocdepth}{2}}
\newpage

\section{Summary Statistics}
\label{sec:appendix_stats}

\subsection{Statistics for Within-Structure Scoring (logSL)}
\label{sec:appendix_stats_logsl}

We provide the full specification of the logSL summary 
statistic vector introduced in Section~\ref{sec:stats}. 
The logSL summary statistic vector is designed to be 
compact and interpretable, capturing the key 
distributional, temporal, and cross-biomarker properties 
of the observed population while remaining computationally 
efficient. It comprises the following four categories:

\begin{enumerate}

\item \textbf{Marginal Distribution Statistics.}
For each biomarker, all simulated values across all 
patients and time points are pooled and standardized 
using the mean and standard deviation of the observed 
data:
$$z = \frac{y - \mu_{\text{obs}}}{\sigma_{\text{obs}}}.$$
Let $\{z_1, z_2, \ldots, z_n\}$ denote the pooled 
standardized values. After sorting,
$$z_{(1)} \le z_{(2)} \le \cdots \le z_{(n)}, 
\qquad q_j = \frac{j}{n+1}, \; j = 1,\dots,n,$$
we approximate the empirical quantile function using 
a first-degree polynomial:
$$z_{(j)} \approx \beta_0 + \beta_1 q_j.$$
The two coefficients $(\beta_0, \beta_1)$ are included 
per biomarker, where $\beta_0$ captures location shifts 
relative to the observed distribution and $\beta_1$ 
captures scale and tail spread. This yields $2p$ 
statistics in total.

\item \textbf{Temporal Dependence Statistics.}
We compute lag-1 Spearman autocorrelation for each 
biomarker. Pooling all valid transitions across 
patients:
$$\rho_{1} = \mathrm{Corr}(y_t,\; y_{t-1}).$$
One autocorrelation coefficient is included per 
biomarker, yielding $p$ statistics in total.

\item \textbf{Population-level Trend Statistics.}
For each biomarker, we compute the Spearman 
correlation between time and the population mean 
trajectory, capturing the overall directional trend 
of the biomarker over the observation window. One 
trend statistic is included per biomarker, yielding 
$p$ statistics in total.

\item \textbf{Dynamic Cross-biomarker Statistics.}
For biomarkers $X$ and $Y$, first differences are 
defined as:
$$\Delta X_t = X_t - X_{t-1}, 
\qquad \Delta Y_t = Y_t - Y_{t-1}.$$
Pooling all valid first differences across patients, 
we compute the Spearman correlation:
$$\rho_{\Delta X,\, \Delta Y} = 
\mathrm{Corr}(\Delta X_t,\; \Delta Y_t).$$
One correlation is included for each unordered 
biomarker pair, yielding $\frac{p(p-1)}{2}$ 
statistics in total.

\end{enumerate}

For $p$ biomarkers, the resulting summary statistic 
vector has dimension $4p + \frac{p(p-1)}{2}$.

\subsection{Statistics for Across-Structure Scoring and 
Diagnosis (MNSD)}
\label{sec:appendix_stats_mnsd}

We list the full set of statistics used for structure comparison via MNSD and as the diagnostic toolkit for the parameter inference agent (Section~\ref{sec:agent}). 
Statistics with empirical IQR $< 0.01$ across observed trajectories are excluded from MNSD computation to avoid division by near-zero values.

\textbf{Per-variable statistics (24 per biomarker):}
\begin{itemize}
    \item \textbf{Distribution (6):} mean, standard 
    deviation, interquartile range, skewness, outlier 
    fraction, value range.
    \item \textbf{Autocorrelation (6):} lag-1, lag-2, 
    lag-3 autocorrelation, first zero-crossing of ACF, 
    dominant frequency, spectral entropy.
    \item \textbf{Stationarity (2):} window mean 
    variation, standard deviation of first differences.
    \item \textbf{Entropy and Complexity (2):} 
    permutation entropy ($m=3$), Lempel-Ziv complexity.
    \item \textbf{Scaling (2):} detrended fluctuation 
    analysis exponent, spectral slope.
    \item \textbf{Model Fit (3):} AR(1) coefficient, 
    exponential smoothing alpha, turning point rate.
    \item \textbf{Trend (1):} Spearman correlation of 
    values versus time.
    \item \textbf{Volatility (2):} mean crossing rate, 
    mean absolute first difference.
\end{itemize}

\textbf{Cross-variable statistics:}
\begin{itemize}
    \item \textbf{Level correlation:} Spearman 
    correlation of levels between two variables.
    \item \textbf{Difference correlation:} Spearman 
    correlation of first differences.
    \item \textbf{Cross-correlation:} cross-correlation 
    at lag 1.
\end{itemize}

\section{Baseline Implementation Details}
\label{sec:appendix_baselines}

We compare AgentODE against baselines encompassing a diverse 
range of methodologies, from sparse regression and evolutionary 
symbolic regression to LLM-guided equation discovery and neural 
network-based approaches. Implementation details for each 
baseline are provided below.

\textbf{SINDy}\quad
SINDy~\citep{brunton2016discovering} is a sparse regression 
method that identifies governing equations by representing 
the time derivatives of state variables as a sparse linear 
combination of candidate library functions. We use the PySINDy implementation~\citep{de2020pysindy}\footnote{\url{https://github.com/dynamicslab/pysindy}} 
with a combined function library of polynomial terms up to 
degree 2 and nonlinear functions including exponential, 
logarithmic, trigonometric, and rational terms. Derivatives 
are estimated using smoothed finite differences, and sparse 
regression is performed using the Sequentially Thresholded 
Least Squares (STLSQ) algorithm. The sparsity threshold is 
tuned per system: $\lambda = 0.02$ for Polymer DA and 
Apoptosis, and $\lambda = 0.20$ for PKPD-Immune.

\textbf{PySR}\quad
PySR~\citep{cranmer2023interpretable} is an evolutionary symbolic regression method based on asynchronous multi-island genetic programming. We use the PySR package\footnote{\url{https://github.com/MilesCranmer/PySR}} and apply it by fitting one symbolic expression per state variable to model its time derivative, estimated from trajectories using Savitzky--Golay smoothing followed by finite differences. The resulting expressions are combined into a full ODE system and integrated forward for evaluation. We use 15 populations of size 33, with 
1000 iterations per variable, and allow binary operators 
$\{+, -, \times, \div\}$ and unary operators 
$\{\exp, \log, \sqrt{\cdot}, \text{square}\}$ with a maximum 
expression size of 20.

\textbf{LLM-SR$^\dagger$}\quad
We use the official implementation of LLM-SR~\citep{shojaee2024llm}\footnote{\url{https://github.com/deep-symbolic-mathematics/LLM-SR}} and extend it from symbolic regression to ODE structure discovery. Specifically, the LLM proposes candidate ODE structures, which are evaluated by simulating synthetic trajectories using numerical integration (via \texttt{odeint}). For each proposed structure, parameters are fitted independently for each trajectory using BFGS to minimize trajectory-level discrepancies. This provides a reference setting with full access to individual-level data.

\textbf{Neural ODE}\quad
Neural ODE~\citep{chen2018neural} learns system dynamics by parameterizing the right-hand side of an ODE with a neural network, $\dot{x} = f_\theta(x)$, and integrating it using a differentiable ODE solver. We use the \texttt{torchdiffeq} library\footnote{\url{https://github.com/rtqichen/torchdiffeq}} with an MLP consisting of 3 hidden layers of size 64 and \texttt{tanh} activations. The model is trained using the Adam optimizer~\citep{kingma2014adam} with learning rate $10^{-3}$ and batch size 32 for up to 500 epochs with early stopping (patience 50) and learning rate decay. Missing values are excluded from the loss computation.

\textbf{LLM-based parameter initialization ($k=5$).}\quad
This ablation independently queries the LLM $k=5$ times with the ODE structure and problem context to propose parameter distributions, without visual summaries and without any diagnosis--update refinement. The parameter distributions achieving the highest logSL score are selected. This ablation serves as a lower bound, reflecting performance without any data-driven iterative refinement.

\textbf{AgentODE w/o iterative refinement.}\quad
This ablation initializes parameter distributions using the parameter initialization prompt with visual summaries of the empirical data, identical to the full AgentODE framework. However, the diagnosis--update loop is disabled and no iterative refinement is performed after initialization. The resulting parameter distributions are used directly to score the candidate ODE structure via MNSD.

\section{AgentODE Implementation Details}
\label{sec:appendix_implementation}

\textbf{LLM Backbone and Compute}\quad
AgentODE uses GPT-5.2 as the backbone LLM for both structure discovery and parameter inference, queried with \texttt{max\_tokens = 8192}. Temperature and top-p are not explicitly set. The primary computational cost comes from LLM API calls; ODE simulation supports both CPU and GPU execution and requires no specialized hardware.

\textbf{Structure Discovery}\quad
The outer loop maintains 10 islands~\citep{shojaee2024llm}, where each island independently evolves a population of candidate ODE structures to promote diversity, with periodic exchange of high-performing structures across islands.
Each LLM call generates 4 candidate structures, with 2 high-scoring structures fetched from the experience buffer to guide each proposal.
The buffer stores all structures grouped into clusters by $-$MNSD score. Clusters are sampled via softmax over Z-scored cluster scores, with temperature decaying from 0.1 toward 0.05 over 30,000 programs, favoring exploitation of high-scoring structures while maintaining diversity.
Island resets occur every 4 hours, wiping the weakest half of islands and seeding each with the best structure from a surviving island.
A structure is registered in the buffer only if parameter inference yields at least one finite logSL, and is scored by $-$MNSD (higher is better) using a single simulation of 1,000 trajectories.

\textbf{logSL Computation}\quad
The log synthetic likelihood is estimated using $N_{\text{sim}} = 150$ simulation replicates, each integrating 100 synthetic trajectories. A trajectory is considered valid if all state variables are finite and fall within predefined physically plausible ranges. A replicate is discarded if fewer than 80\% of trajectories are valid, in which case logSL $\to -\infty$. A small regularization term $\lambda = 10^{-6}$ is added to the diagonal of $\hat{\Sigma}_\psi$ before inversion to ensure numerical stability.

\textbf{Constraint violation handling.}\quad
When no parameter distributions proposed so far have yielded a finite logSL, the loop enters constraint violation mode. A violation report is generated for the current parameter distributions, describing which state variables are out of physically plausible range and the fraction of invalid trajectories. A dedicated \texttt{CONSTRAINT\_VIOLATION} prompt is constructed with the current parameters and violation report, and sent to the LLM to propose corrected distributions. 

\textbf{Parameter Inference}\quad
The inner diagnosis--update loop runs for a default of 10 iterations. If the candidate structure yields a finite logSL and its $-$MNSD score exceeds the island's current best, the loop is extended to 20 iterations, allowing more thorough parameter inference for promising structures without committing extra computation to poor ones. 
Early stopping applies a patience of 5, counting both non-improving steps and consecutive constraint-violating steps. The best iteration is selected by the highest logSL. 
The diagnosis phase spans up to 5 tool-calling turns, after which an update prompt is constructed with the 3 best parameter sets by logSL from the current iteration log.
\section{Benchmark Details}
\label{sec:appendix_benchmarks}

Standard benchmarks such as Lotka-Volterra and SIR 
are well-known to be susceptible to LLM recitation. 
We therefore selected domain-specific systems and 
verified their resistance to memorization through 
the contamination analysis in 
Table~\ref{tab:contamination} using the prompt 
provided in Appendix~\ref{sec:appendix_prompts_contamination}.

\begin{table}[t]
\centering
\caption{Contamination analysis for benchmark systems. Recovery counts exact structural fingerprint matches across 5 runs using \texttt{GPT-5.2}.}
\label{tab:contamination}
\small
\begin{tabular}{
>{\raggedright\arraybackslash}p{2.2cm}
>{\raggedright\arraybackslash}p{2.9cm}
>{\raggedright\arraybackslash}p{5.2cm}
>{\centering\arraybackslash}p{1.2cm}
}
\toprule
\textbf{Benchmark} &
\textbf{Variables} &
\textbf{Key structural features} &
\textbf{Recovery} \\
\midrule
Apoptosis
& Initiator, regulator, effector
& Multiplicative feedback
$x_{\mathrm{effector}}(x_{\mathrm{regulator}}+0.1)$; conservation law
& \textbf{0/5} \\
\addlinespace
Polymer DA
& Diene, dienophile, adduct
& Dual bilinear terms ($x_0x_1$, $x_0x_3$); $x_3$ acts as reactant
& \textbf{0/5} \\
\addlinespace
PKPD-Immune
& Depot, central, immune, effector
& Saturating interaction
$\frac{\rho\,x_{\mathrm{immune}}\,x_{\mathrm{effector}}}
{x_{\mathrm{immune}}+\sigma}$; quadratic effector term
& \textbf{0/5} \\
\bottomrule
\end{tabular}
\end{table}

We provide the full ODE specifications, parameter 
settings, and data generation details for each 
benchmark below.

\subsection{Apoptosis}
\label{sec:appendix_apoptosis}
The apoptosis system models programmed cell death through the 
interactions of three coupled protein concentrations: initiator 
protein, regulator protein, and effector protein. The governing 
ODE system is adapted from~\citet{bideh2026llm}:
\begin{align*}
\frac{d\,\text{initiator}}{dt} &= -\frac{k_1 \cdot \text{initiator} \cdot \text{regulator}}{\text{initiator} + s_1} - k_2 \cdot \text{initiator} + k_3 \\[6pt]
\frac{d\,\text{regulator}}{dt} &= -\frac{k_4 \cdot \text{initiator} \cdot \text{regulator}}{\text{regulator} + s_2} - \frac{k_5 \cdot \text{regulator}}{\text{regulator} + s_3} + k_6 \cdot \text{effector} \cdot (\text{regulator} + s_3) \\[6pt]
\frac{d\,\text{effector}}{dt} &= \phantom{-}\frac{k_4 \cdot \text{initiator} \cdot \text{regulator}}{\text{regulator} + s_2} + \frac{k_5 \cdot \text{regulator}}{\text{regulator} + s_3} - k_6 \cdot \text{effector} \cdot (\text{regulator} + s_3)
\end{align*}
with fixed parameters $k_1 = 0.40$, $k_2 = 0.05$, $k_3 = 0.10$, 
$k_4 = 7.95$, $k_5 = 0.20$, $k_6 = 0.60$, $s_1 = 0.10$, 
$s_2 = 2.00$, and $s_3 = 0.10$. All parameters are fixed across 
individuals; only initial conditions vary.

\textbf{Data Generation.}\quad
A cohort of 100 trajectories is simulated over $t \in [0, 10]$ 
with 101 evenly spaced time points ($\Delta t = 0.1$). Initial 
conditions are sampled independently from uniform distributions:
initiator $\sim \mathcal{U}(0.001, 0.05)$,
regulator $\sim \mathcal{U}(0.05, 0.50)$,
effector $\sim \mathcal{U}(0.50, 3.00)$.
The cohort is split into 80 inference trajectories and 20 test 
trajectories drawn from the same initial condition distributions.

\subsection{Polymer DA Cross-linking}
\label{sec:appendix_polymer}

The Diels-Alder (DA) cross-linking system models the reversible 
covalent reaction between a diene and a dienophile component, 
a thermally reversible process widely studied in polymer chemistry 
for self-healing and recyclable materials. The system involves four 
coupled state variables: diene concentration ($x_0$), dienophile 
concentration ($x_1$), cross-linked adduct concentration ($x_2$), 
and de-cross-linked component concentration ($x_3$). The governing 
ODE system is adapted from~\citet{bideh2026llm}:

\begin{align*}
\frac{d\,x_0}{dt} &= -k_1 x_0 x_1 - k_2 x_0 x_3 + k_3 x_1 + k_4 x_2 \\
\frac{d\,x_1}{dt} &= -k_1 x_0 x_1 + k_2 x_0 x_3 - k_3 x_1 + k_4 x_2 \\
\frac{d\,x_2}{dt} &=  k_1 x_0 x_1 - k_4 x_2 \\
\frac{d\,x_3}{dt} &= -k_2 x_0 x_3 + k_3 x_1
\end{align*}

with fixed parameters $k_1 = 0.03$, $k_2 = 0.01$, $k_3 = 0.08$, 
and $k_4 = 0.03$. All rate constants are fixed across individuals; 
only initial conditions vary.

\textbf{Data Generation.}\quad
The original system from~\citet{bideh2026llm} uses fixed initial 
conditions. To simulate a heterogeneous population suitable for 
summary-level inference, we extend the initial conditions to be 
sampled independently from uniform distributions:
$x_0 \sim \mathcal{U}(1.0, 30.0)$,
$x_1 \sim \mathcal{U}(0.0, 0.1)$,
$x_2 \sim \mathcal{U}(0.0, 0.1)$,
$x_3 \sim \mathcal{U}(0.5, 10.0)$,
A cohort of 100 trajectories is simulated over $t \in [0, 50]$ 
with 101 evenly spaced time points ($\Delta t = 0.5$). The cohort is 
split into 80 inference trajectories and 20 test trajectories 
drawn from the same initial condition distributions.

\subsection{PKPD-Immune}
\label{sec:appendix_pkpd}
The PKPD-Immune system is a composite four-variable ODE coupling 
two-compartment pharmacokinetics with an immune suppression and 
effector cell response module. The state variables are: drug amount 
in the absorption compartment (depot), drug amount in the central 
compartment (central), immune cell population (immune), and 
effector cell population (effector). The governing ODE system is:
\begin{align*}
\frac{d\,\text{depot}}{dt} &= -k_a \cdot \text{depot} \\[6pt]
\frac{d\,\text{central}}{dt} &= k_a \cdot \text{depot} - k_e \cdot \text{central} \\[6pt]
\frac{d\,\text{immune}}{dt} &= s - d \cdot \text{immune} - \frac{k_1 (\text{central}/V) \cdot \text{immune}}{IC_{50} + \text{central}/V} \\[6pt]
\frac{d\,\text{effector}}{dt} &= \frac{\rho \cdot \text{immune} \cdot \text{effector}}{\text{immune} + \sigma} - \delta \cdot \text{effector} - \varepsilon \cdot \text{effector}^2
\end{align*}
where $\varepsilon = 0.05$ is a fixed structural parameter governing 
effector intraspecific competition. Both initial conditions and model 
parameters vary across individuals, reflecting pharmacokinetic and 
pharmacodynamic variability in drug response. All parameters are 
sampled independently from log-normal distributions; 
Table~\ref{tab:pkpd_params} reports the mean and standard deviation 
in linear space.
\begin{table}[h]
\centering
\small
\begin{tabular}{llcc}
\toprule
\textbf{Parameter} & \textbf{Description} & \textbf{Mean} & \textbf{SD} \\
\midrule
$k_a$      & Absorption rate (day$^{-1}$)           & 1.00 & 0.25 \\
$k_e$      & Elimination rate (day$^{-1}$)           & 0.30 & 0.08 \\
$V$        & Volume of distribution                  & 10.0 & 2.00 \\
$s$        & Immune cell source rate                 & 1.00 & 0.20 \\
$d$        & Immune cell death rate                  & 0.15 & 0.04 \\
$k_1$      & Drug suppression rate                   & 0.80 & 0.20 \\
$IC_{50}$  & Half-suppression concentration          & 0.50 & 0.15 \\
$\rho$     & Effector recruitment rate               & 0.40 & 0.10 \\
$\sigma$   & Recruitment half-saturation             & 2.00 & 0.50 \\
$\delta$   & Effector death rate                     & 0.20 & 0.05 \\
\bottomrule
\end{tabular}
\caption{PKPD-Immune parameter distributions (log-normal).}
\label{tab:pkpd_params}
\end{table}

\textbf{Data Generation.}\quad
A cohort of 100 trajectories is simulated over $t \in [0, 50]$ days 
with 101 evenly spaced time points ($\Delta t = 0.5$ days). 
Initial conditions are sampled from log-normal distributions: 
depot $\sim \text{LogNormal}(\mu=5.0, \sigma=1.0)$, 
central $= 0$ (central compartment always empty at dose time), 
immune $\sim \text{LogNormal}(\mu=6.67, \sigma=1.5)$ initialized 
near the drug-free immune steady state, and 
effector $\sim \text{LogNormal}(\mu=1.0, \sigma=0.30)$. The cohort 
is split into 80 inference trajectories and 20 test trajectories 
drawn from the same distributions.

\section{Clinical Dataset Details}
\label{sec:appendix_clinical}

\subsection{Acute Kidney Injury (AKI)}
\label{sec:appendix_aki}

\textbf{Patient Selection.}\quad
Patients aged 50--64 years at admission are selected from MIMIC-IV 3.1~\citep{johnson2023mimic}\footnote{\url{https://physionet.org/content/mimiciv/3.1/}}. 
AKI is 
confirmed using ICD-10 diagnosis codes (N17.*) and/or KDIGO 
clinical criteria ($\geq$0.3~mg/dL creatinine increase within 
48 hours, or $\geq$1.5$\times$ baseline within 7 days). Only 
admissions with at least 20 unique timepoints within the first 
168 hours and at least two of the three core biomarkers present 
at each timepoint are retained. The final cohort comprises 345 
patients (353 admissions, 8,249 timepoints; mean age 
$57.4 \pm 4.4$ years).

\textbf{Train/Test Split.}\quad
The cohort is split at the admission level into 80\% inference 
(282 patients) and 20\% test (71 patients).

\textbf{Preprocessing.}\quad
No interpolation or imputation is applied. Missing values and 
irregular sampling are handled natively by AgentODE through 
population-level summary statistics, and by Neural ODE and 
LLM-SR-ODE through masked loss computation.

\textbf{Ethics.}\quad
MIMIC-IV is accessed under the PhysioNet Credentialed Health 
Data License 1.5.0 and the PhysioNet Credentialed Health Data 
Use Agreement 1.5.0. No individual-level data are released as 
part of this work.

\textbf{Preprocessing.}\quad
No interpolation or imputation is applied to the raw measurements. 
Missing values and irregular sampling are handled natively by 
AgentODE through population-level summary statistics, and by 
Neural ODE and LLM-SR-ODE through masked loss computation.

\subsection{Recessive Dystrophic Epidermolysis Bullosa (RDEB)}
\label{sec:appendix_rdeb}

\textbf{Dataset.}\quad
The dataset comprises 46 patients with severe 
RDEB~\citep{bardhan2020epidermolysis, reimer2020natural, 
karakioulaki2026systemic}. Patients ranged from 0 to 30.4 years 
of age at their first recorded visit. 
We focus on longitudinal follow-up spanning $t \in [0, 144]$ 
months since first hospital visit (mean 53.3 months per patient). A total of 231 observations 
are available across 46 patients, with a mean of 5.0 timepoints 
per patient and 16 patients having as few as 2 timepoints.

\textbf{State Variables.}\quad
Four state variables are modeled: serum albumin (g/dL), 
log-transformed C-reactive protein 
(log$_{10}$(CRP$+$1), where CRP is in mg/L), hemoglobin 
(g/dL), and BMI relative to age- and sex-matched WHO reference 
(BMI\textsubscript{rel}, kg/m$^2$). BMI\textsubscript{rel} 
is computed as the deviation of each patient's BMI from the 
age- and sex-matched WHO healthy reference, such that a value 
of 0 indicates a healthy BMI and negative values indicate 
underweight relative to peers.

\textbf{Preprocessing.}\quad
CRP is log$_{10}$(CRP$+$1) transformed to stabilize variance. 
BMI\textsubscript{rel} is computed as the deviation from 
age- and sex-matched WHO reference values. No interpolation 
or imputation is applied.

\textbf{Evaluation.}\quad
Due to the limited cohort size, no train/test split is performed 
and results are reported in-sample, as noted in 
Table~\ref{tab:results}.

\textbf{Ethics.}\quad
The RDEB dataset was collected under institutional ethics approval 
at the authors' institution. All patients or their legal 
guardians provided informed consent. No individual-level data 
are released as part of this work.
\section{Additional Results}
\label{sec:appendix_results}

\subsection{Apoptosis}
\label{sec:appendix_ode_apop}

\begin{figure}[p]
    \centering
    \includegraphics[
        height=0.92\textheight,
        keepaspectratio
    ]{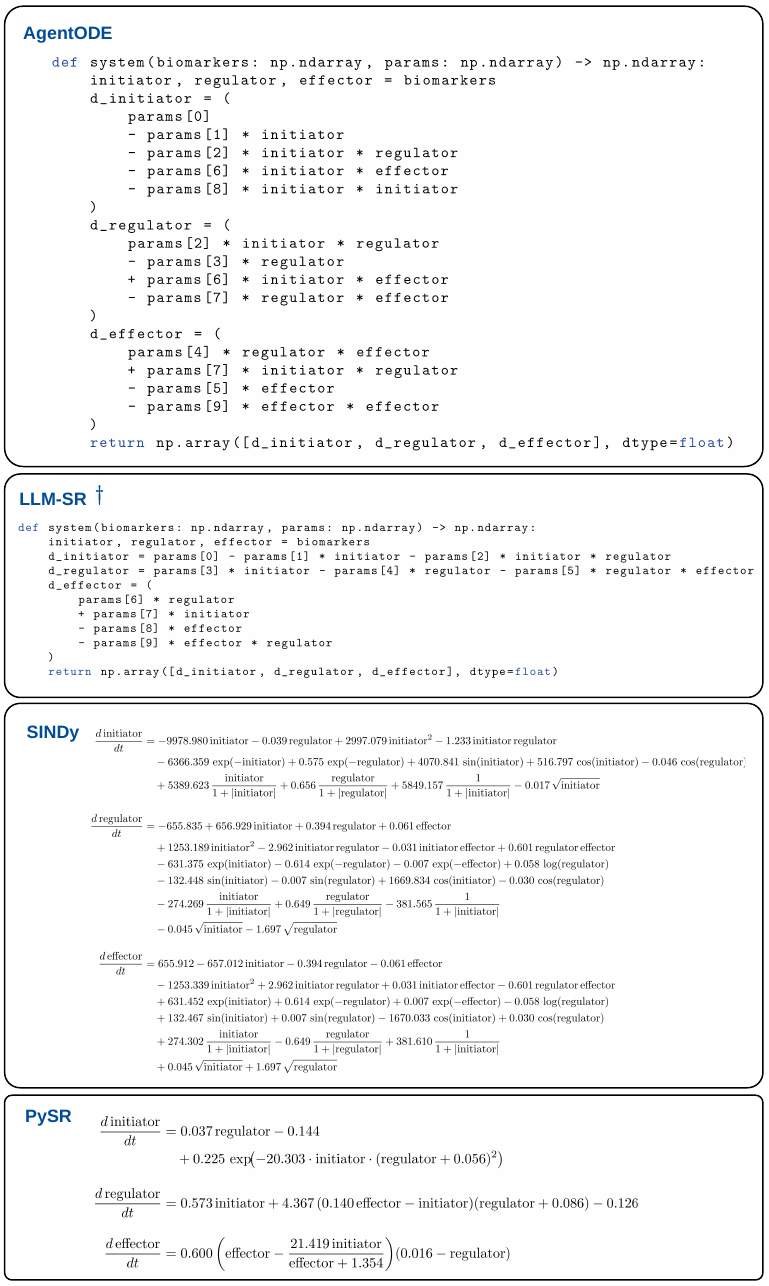}
    \caption{Recovered ODE structures for the Apoptosis benchmark using AgentODE, LLM-SR$^\dagger$, SINDy, and PySR.}
    \label{fig:apop_structures}
\end{figure}

Figure~\ref{fig:apop_structures} compares the ODE structures recovered by different methods on the Apoptosis benchmark. 
SINDy recovers highly complex equations containing many redundant nonlinear, trigonometric, and exponential terms, producing expressions that are difficult to interpret mechanistically despite achieving strong numerical performance. PySR discovers substantially more compact equations, but the recovered interactions deviate from the underlying system structure and fail to preserve the coupled regulatory dynamics between variables.
LLM-SR$^\dagger$ recovers a sparse and biologically interpretable structure that closely resembles the ground-truth interaction pattern, correctly identifying key cross-variable couplings between initiator, regulator, and effector dynamics. AgentODE similarly recovers a compact interaction structure. Although the exact symbolic form differs from the ground truth, the discovered system preserves the core dynamical relationships between variables, including inhibitory and recruitment-like interactions that govern the population-level apoptosis behavior.

\subsection{Polymer DA Cross-linking}
\label{sec:appendix_ode_polymer}

\begin{figure}[p]
    \centering
    \includegraphics[
        height=0.92\textheight,
        keepaspectratio
    ]{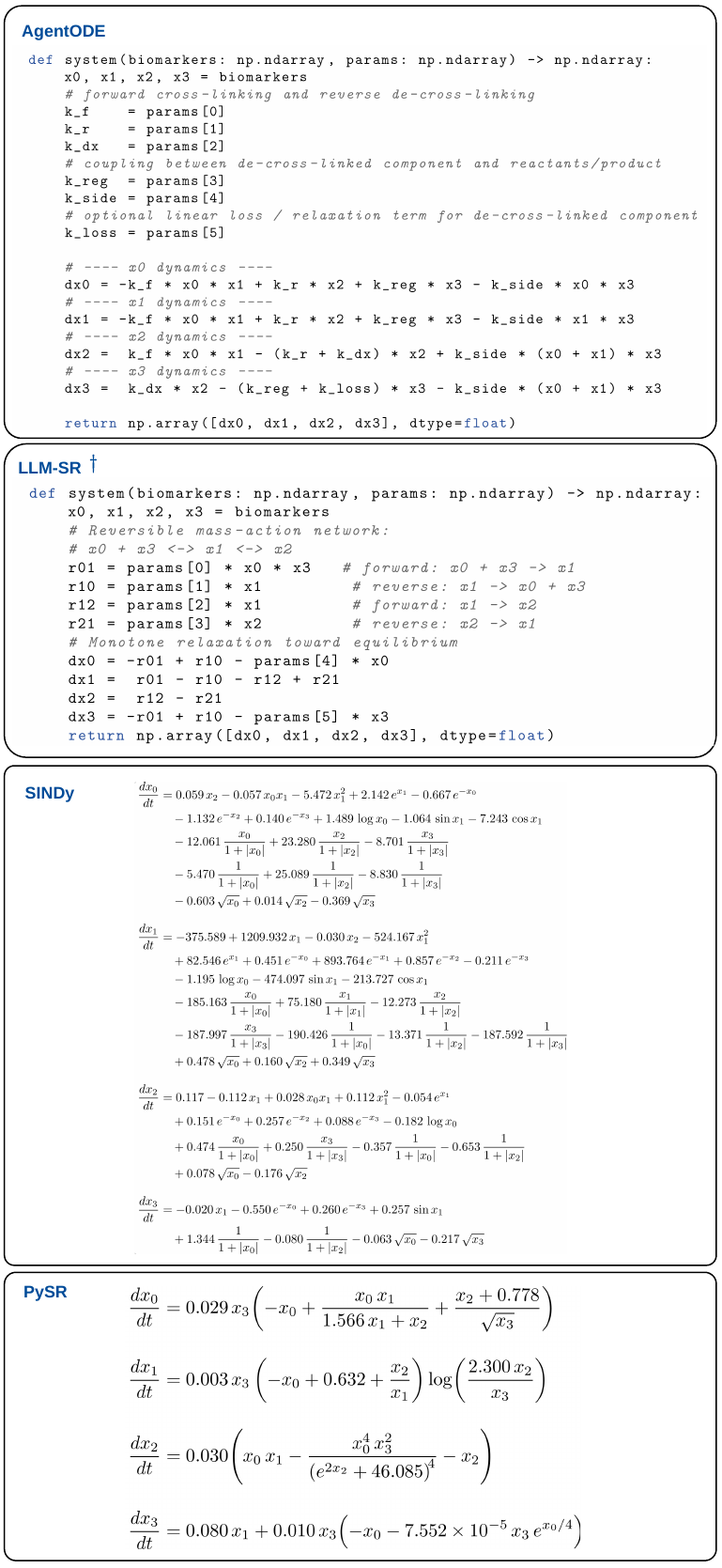}
    \caption{Recovered ODE structures for the Polymer DA Cross-linking benchmark using AgentODE, LLM-SR$^\dagger$, SINDy, and PySR.}
    \label{fig:polymer_structures}
\end{figure}

Figure~\ref{fig:polymer_structures} compares the ODE structures recovered by different methods on the Polymer DA Cross-linking benchmark. The ground-truth system is governed by sparse reaction-like terms involving cross-linking between reactants, product formation, and de-cross-linking dynamics.
SINDy again recovers highly complex equations with many nonlinear, exponential, logarithmic, and trigonometric terms, making the resulting system difficult to interpret mechanistically. PySR produces more compact equations, but the recovered expressions contain nonphysical functional forms such as logarithmic and high-order rational terms, which do not align well with the underlying reaction kinetics.
LLM-SR$^\dagger$ recovers an interpretable reversible reaction network, but its proposed structure differs from the ground-truth reaction topology. In contrast, AgentODE recovers a compact mass-action-like system with forward cross-linking, reverse de-cross-linking, and side-interaction terms. Although the exact coefficients and terms differ from the ground truth, the recovered structure preserves the main mechanistic motifs of the Polymer system, including reactant-product conversion and coupling through the de-cross-linked component.

\subsection{PKPD-Immune}
\label{sec:appendix_ode_pkpd}

\begin{figure}[p]
    \centering
    \includegraphics[
        height=0.92\textheight,
        keepaspectratio
    ]{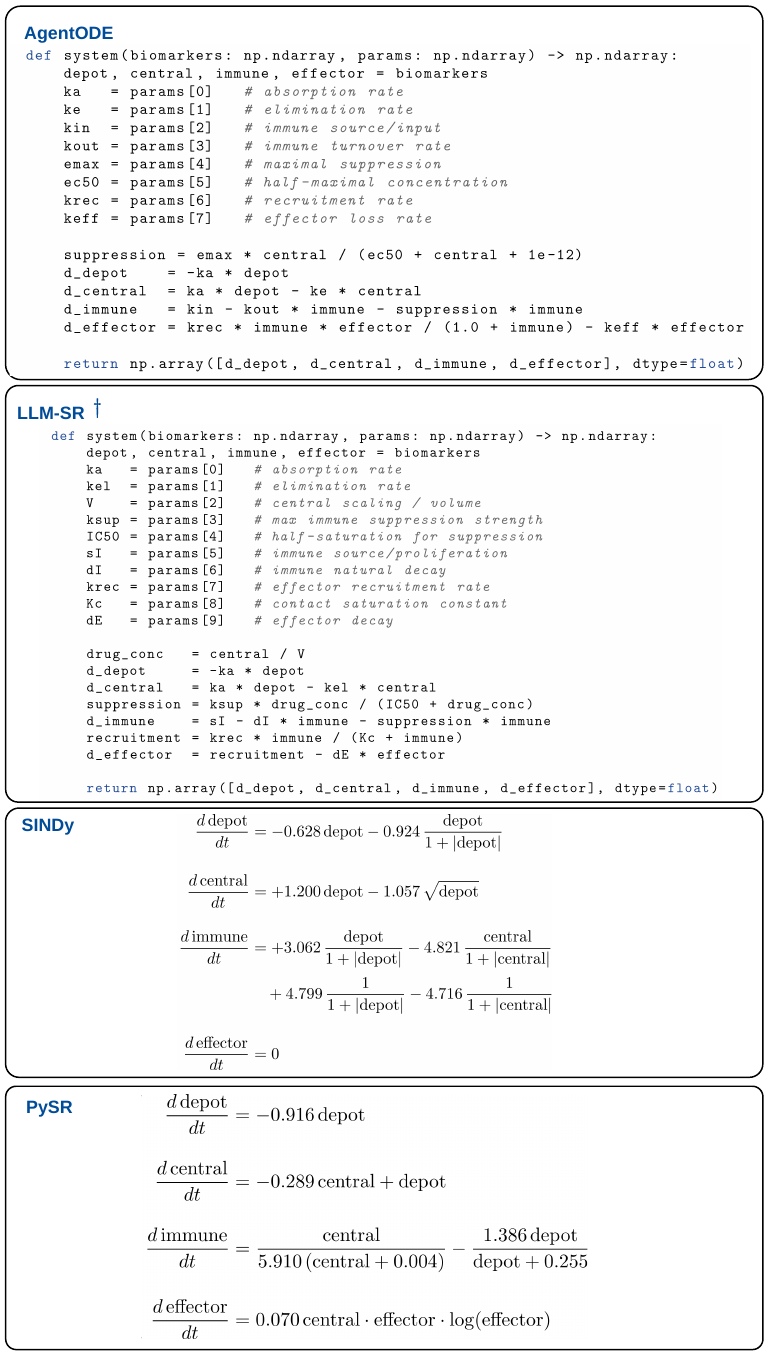}
    \caption{Recovered ODE structures for the PKPD-Immune benchmark using AgentODE, LLM-SR$^\dagger$, SINDy, and PySR.}
    \label{fig:pkpd_structures}
\end{figure}

Figure~\ref{fig:pkpd_structures} compares the ODE structures recovered by different methods on the PKPD-Immune benchmark. The ground-truth system combines two-compartment pharmacokinetics with immune suppression and effector-cell recruitment dynamics.
SINDy recovers a simplified structure that captures parts of the depot and central compartment dynamics, but fails to model effector dynamics and omits key immune--effector interactions. PySR produces compact equations but introduces nonphysical terms, including logarithmic effector feedback, and does not recover the mechanistic immune recruitment structure.
In contrast, both LLM-SR$^\dagger$ and AgentODE recover interpretable PKPD-style systems. LLM-SR$^\dagger$ closely matches the ground-truth structure, including absorption, elimination, saturable immune suppression, and effector decay. AgentODE recovers the same core pharmacokinetic and immune-suppression mechanisms while using only summary-level statistics. Although its effector recruitment term differs from the ground truth, the recovered structure preserves the main causal pathway from depot absorption to central drug exposure, immune suppression, and effector dynamics.

\subsection{AKI}
\label{sec:appendix_ode_aki}

\paragraph{ODE structure discovered by AgentODE.}

AgentODE discovers a linear coupled biomarker system:

\begin{align*}
\frac{d\mathrm{Cr}}{dt} &= \theta_0 - \theta_1 \mathrm{Cr} + \theta_2 \mathrm{BUN}, \\
\frac{d\mathrm{BUN}}{dt} &= \theta_3 - \theta_4 \mathrm{BUN} + \theta_5 \mathrm{Cr}, \\
\frac{d\mathrm{K}}{dt} &= \theta_6 - \theta_7 \mathrm{K} + \theta_8 \mathrm{Cr} + \theta_9 \mathrm{BUN}.
\end{align*}

This structure represents each biomarker through a baseline source term, a clearance or recovery term, and cross-biomarker coupling. Creatinine and BUN are mutually coupled, reflecting shared renal dysfunction and azotemia dynamics, while potassium is driven by both creatinine and BUN together with its own elimination term. The resulting system is parsimonious and clinically interpretable.

\paragraph{ODE structure discovered by LLM-SR$^\dagger$.}
LLM-SR$^\dagger$ discovers the following AKI system:

\begin{align*}
\frac{d\mathrm{Cr}}{dt} &= \theta_0 + \theta_1 \mathrm{Cr} - \theta_2 \mathrm{Cr}, \\
\frac{d\mathrm{BUN}}{dt} &= \theta_3 + \theta_4 \mathrm{Cr} + \theta_5 \mathrm{BUN} - \theta_6 \mathrm{BUN}, \\
\frac{d\mathrm{K}}{dt} &= \theta_7 \mathrm{Cr} + \theta_8 \mathrm{BUN} + \theta_9 - \theta_{10} \mathrm{K}.
\end{align*}

This structure also captures clinically plausible AKI dynamics, including creatinine-driven BUN changes and potassium dependence on renal dysfunction markers. However, the positive and negative self-effects for creatinine and BUN reduce to net linear turnover terms, making these effects harder to distinguish under sparse clinical sampling. Compared with AgentODE, the LLM-SR$^\dagger$ structure is slightly less parsimonious and lacks direct BUN-to-creatinine coupling.

\subsection{RDEB}
\label{sec:appendix_ode_rdeb}

\paragraph{ODE structure developed independently by ODE experts in collaboration with RDEB clinicians.}
The expert-derived model uses four state variables: nutrition ($N$), inflammation ($I$), anemia state ($A$), and general health ($H$), along with an unobserved auxiliary inflammatory drive $\widetilde{I}$. These states were mapped to the previously used clinical proxy variables albumin, C-reactive protein, hemoglobin, and BMI relative to age- and sex-matched WHO reference values, respectively. To facilitate quantitative interpretation of model parameters, clinical variables were standardized to a~$[0,1]$ range. The model is specified by:

\begin{align*}
\frac{dN}{dt} &= \lambda_N \left(N^*(I) - N\right), \quad
N^*(I) = \frac{k_{I,N}(1 - I)}{k_{I,N} + I} \\[6pt]
\frac{dI}{dt} &= \lambda_I \left(I^*(H, \widetilde{I}) - I\right), \quad
I^*(H, \widetilde{I}) = \frac{1}{1+\widetilde{I}}\left[\frac{k_{H,I}(1-H)}{k_{H,I}+H} + \widetilde{I}\right] \\[6pt]
\frac{dA}{dt} &= \lambda_A \left(A^*(I) - A\right), \quad
A^*(I) = \frac{k_{I,A}(1 - I)}{k_{I,A} + I} \\[6pt]
\frac{dH}{dt} &= \lambda_H \left(H^*(N, A, I) - H\right), \quad
H^*(N, A, I) = \frac{1}{3}\left(\frac{k_{I,H}(1-I)}{k_{I,H}+I} + \frac{(1+k_{N,H})N}{k_{N,H}+N} + \frac{(1+k_{A,H})A}{k_{A,H}+A}\right) \\[6pt]
\frac{d\widetilde{I}}{dt} &= p_I, \quad \widetilde{I}(0) = 0
\end{align*}

\noindent where each observed state relaxes toward a state-specific set point that depends on the current value of the other states. The auxiliary state $\widetilde{I}$ represents the unobserved progressive inflammatory drive and is not used as an observed clinical variable.

\paragraph{ODE structure discovered by AgentODE.}
The following ODE structure models the coupled dynamics of serum albumin, log-transformed C-reactive protein ($\log$-CRP), hemoglobin, and BMI relative to age- and sex-matched WHO reference (BMI\textsubscript{rel}).

\begin{align*}
\frac{d\, \text{albumin}}{dt} &= \theta_0 \left(\theta_1 - \text{albumin}\right) - \theta_2 \cdot \log\text{CRP} \\[6pt]
\frac{d\, \log\text{CRP}}{dt} &= \theta_3 \left(\theta_4 - \log\text{CRP}\right) + \theta_5 \left(\theta_1 - \text{albumin}\right) \\[6pt]
\frac{d\, \text{Hb}}{dt} &= \theta_6 \left(\theta_7 - \text{Hb}\right) - \theta_8 \cdot \log\text{CRP} + \theta_9 \left(\text{albumin} - \theta_1\right) \\[6pt]
\frac{d\, \text{BMI}_{\text{rel}}}{dt} &= \theta_{10} \left(\theta_{11} - \text{BMI}_{\text{rel}}\right) + \frac{\theta_9}{2} \left(\text{albumin} - \theta_1\right)
\end{align*}

\begin{figure}[h]
    \centering
    \includegraphics[width=1.0\textwidth]{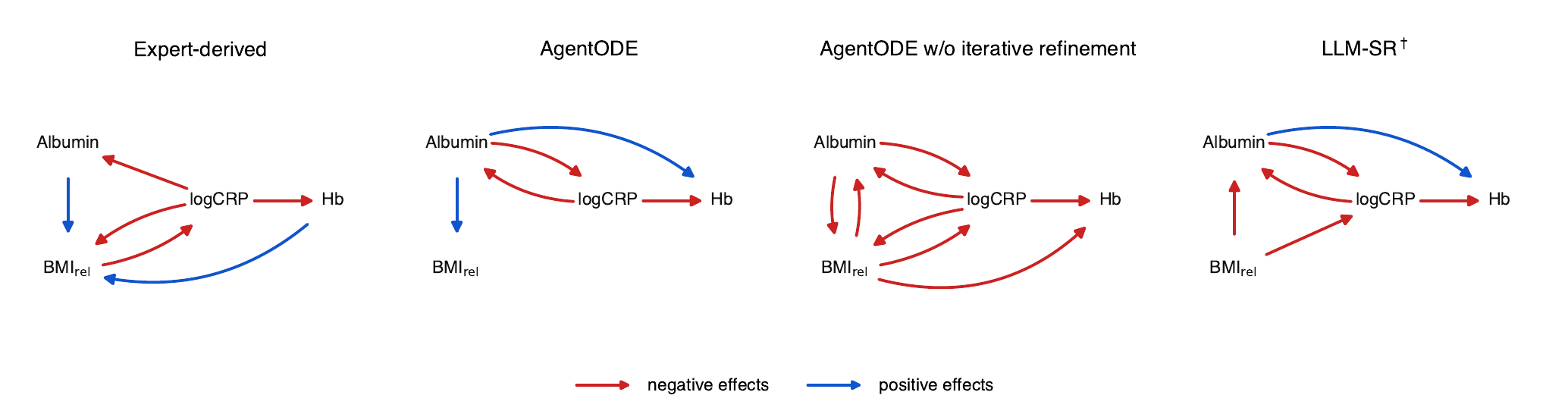}
    \caption{Causal diagrams for the RDEB ODE structures from (\textbf{first}) the expert-derived, (\textbf{second}) AgentODE, (\textbf{third}) AgentODE w/o iterative refinement, and (\textbf{fourth}) LLM-SR$^\dagger$. Red arrows indicate negative effects, blue arrows indicate positive effects.}
    \label{fig:rdeb_causal}
\end{figure}

\textbf{Comparison.}\quad
The expert-derived model and the model found by AgentODE approach the phenomenological relationships between variables differently, but both recognize the key disease loop, as visualized in Figure~\ref{fig:rdeb_causal}: inflammation (log-CRP) drives worsening of the other clinical variables, while a second variable attenuates inflammation. In the expert model, BMI$_{\text{rel}}$ plays this protective role as a proxy for patient health, while AgentODE identifies albumin as the protective variable, interpreted as a proxy for nutritional status. Both models plausibly link the clinical variables into a phenomenological description, for which no unique solution exists.

\paragraph{ODE structure discovered by AgentODE w/o iterative refinement.}
The following structure achieves a marginally lower RMSE (0.628 vs 0.644) than the full AgentODE model. However, mechanistic inspection reveals several implausibilities. The BMI$_{\text{rel}}$ equation reuses parameters $\theta_0, \theta_1, \theta_2, \theta_3$ from the albumin equation, imposing the constraint that both variables share identical relaxation rates and baselines, which is clinically implausible as they reflect distinct physiological processes. More fundamentally, higher BMI$_{\text{rel}}$ negatively affects albumin and hemoglobin, whereas higher BMI$_{\text{rel}}$ generally reflects better nutritional status and should be associated with improved clinical markers. Conversely, higher albumin negatively affects BMI$_{\text{rel}}$, which contradicts the established association between good nutritional status and healthy body weight. As shown in Figure~\ref{fig:rdeb_causal}, the resulting causal structure is dense with exclusively negative cross-variable effects and no clear protective mechanism, in contrast to the parsimonious and clinically interpretable structure recovered by AgentODE. This illustrates that in sparse and noisy clinical settings, lower RMSE may reflect overfitting to noise rather than a more mechanistically valid model.



\begin{align*}
\frac{d\, \text{albumin}}{dt} &= \theta_0 \left(\theta_1 - \text{albumin}\right) - \theta_2 \cdot \log\text{CRP} - \theta_3 \cdot \text{BMI}_{\text{rel}} \\[6pt]
\frac{d\, \log\text{CRP}}{dt} &= \theta_4 \left(\theta_5 - \log\text{CRP}\right) - \theta_6 \cdot \text{albumin} - \theta_7 \cdot \text{BMI}_{\text{rel}} \\[6pt]
\frac{d\, \text{Hb}}{dt} &= \theta_8 \left(\theta_9 - \text{Hb}\right) - \theta_{10} \cdot \log\text{CRP} - \theta_{11} \cdot \text{BMI}_{\text{rel}} \\[6pt]
\frac{d\, \text{BMI}_{\text{rel}}}{dt} &= \theta_0 \left(\theta_1 - \text{BMI}_{\text{rel}}\right) - \theta_2 \cdot \log\text{CRP} - \theta_3 \cdot \text{albumin}
\end{align*}

\paragraph{ODE structure discovered by LLM-SR$^\dagger$.}
This structure achieves a lower RMSE (0.532) than AgentODE (0.644) by treating BMI$_{\text{rel}}$ as a static variable with zero derivative. While this trivially reduces prediction error on a sparse and noisy variable, it entirely forgoes modeling BMI$_{\text{rel}}$ dynamics. Since BMI$_{\text{rel}}$ is a key indicator of disease progression and nutritional status in RDEB, a model that does not capture its temporal evolution cannot provide mechanistic insights into this clinically important dimension. Furthermore, the albumin and log-CRP equations use constant baseline synthesis terms ($\theta_0$ and $\theta_4$) rather than relaxation toward a physiologically meaningful baseline, which limits biological interpretability. As shown in Figure~\ref{fig:rdeb_causal}, BMI$_{\text{rel}}$ is entirely disconnected from the causal structure. This further illustrates that lower RMSE does not necessarily reflect a more mechanistically informative model in sparse and noisy clinical settings.

\begin{align*}
\frac{d\, \text{albumin}}{dt} &= \theta_0 - \theta_1 \cdot \text{albumin} - \theta_2 \cdot \log\text{CRP} - \theta_3 \cdot \text{BMI}_{\text{rel}} \\[6pt]
\frac{d\, \log\text{CRP}}{dt} &= \theta_4 - \theta_5 \cdot \log\text{CRP} - \theta_6 \cdot \text{albumin} - \theta_7 \cdot \text{BMI}_{\text{rel}} \\[6pt]
\frac{d\, \text{Hb}}{dt} &= \theta_8 - \theta_9 \cdot \text{Hb} - \theta_{10} \cdot \log\text{CRP} + \theta_{11} \cdot \text{albumin} \\[6pt]
\frac{d\, \text{BMI}_{\text{rel}}}{dt} &= 0
\end{align*}

\paragraph{Analysis.} 
These comparisons highlight a fundamental advantage of summary-level inference in clinical settings. Methods that optimize directly against individual trajectories, such as LLM-SR$^\dagger$, can achieve low RMSE by overfitting to noise rather than recovering the true underlying mechanism. In contrast, reasoning from population-level summary statistics encourages the LLM to identify patterns that are robust across the cohort, promoting mechanistically grounded structure discovery. This distinction becomes particularly consequential in the experience buffer: since high-scoring structures are retrieved to guide future proposals, a mechanistically flawed structure with artificially low RMSE can mislead the search toward implausible regions of the structure space. Summary statistics, by averaging out individual-level noise, provide a more representative and informative signal for clinical ODE discovery than trajectory-level fitting alone.

\paragraph{Inferred parameter distributions}
\label{sec:param_dist}

\begin{figure}[h]
    \centering
    \includegraphics[width=\textwidth]{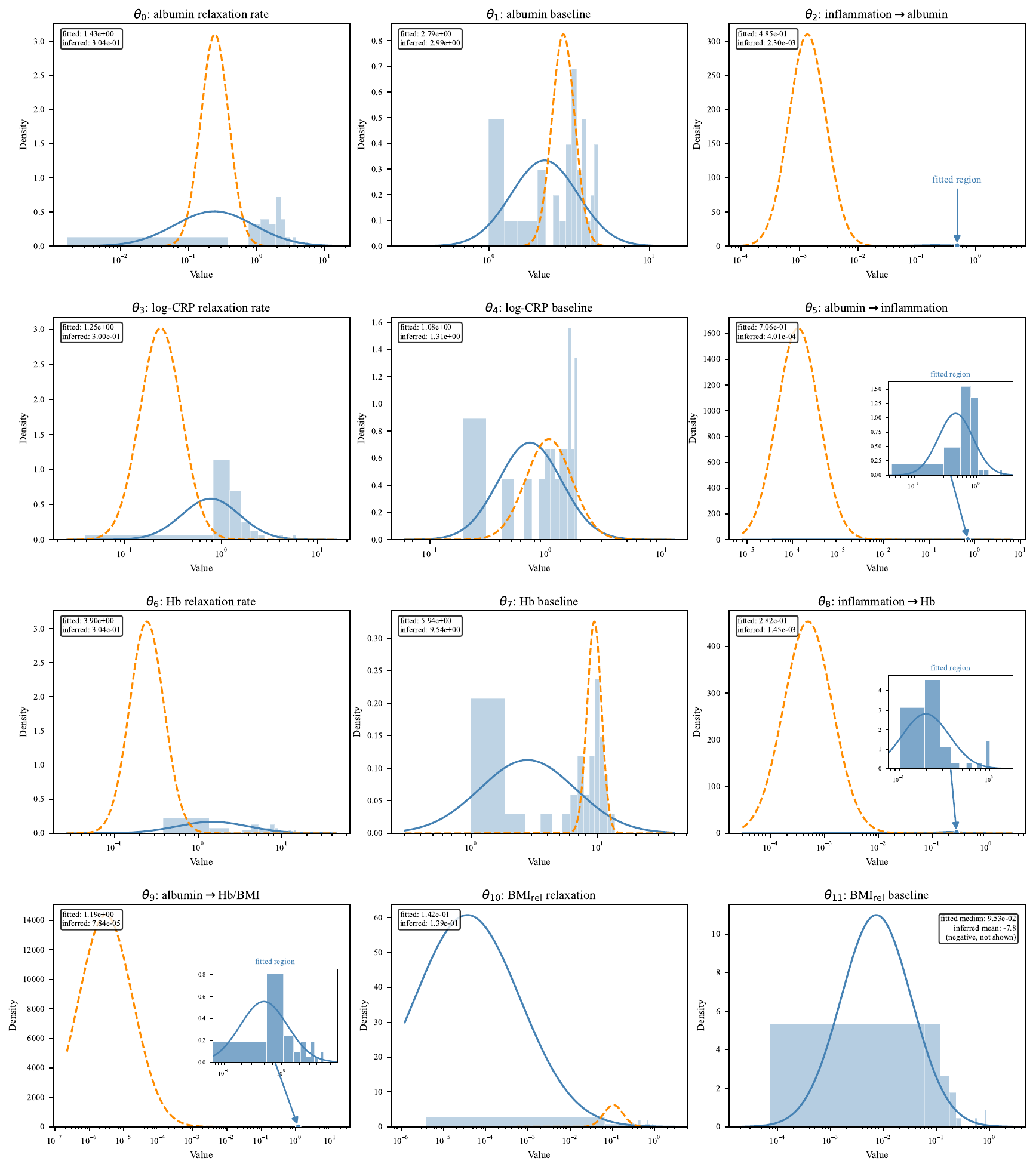}
    \caption{Parameter distribution comparisons for the AgentODE-discovered RDEB ODE structure. \textbf{Blue histograms}: per-patient parameter values from BFGS fitting to individual trajectories. \textbf{Blue solid curves}: log-normal fits to these values. \textbf{Orange dashed curves}: log-normal distributions inferred by AgentODE from population-level summary statistics.}
    \label{fig:param_dist_rdeb}
\end{figure}

Figure~\ref{fig:param_dist_rdeb} compares the parameter distributions inferred by AgentODE against empirical distributions for the RDEB dataset. To obtain the empirical distributions, we fit the discovered ODE structure to each individual trajectory independently via BFGS optimization, yielding one parameter vector per patient. The resulting per-patient parameter values are shown as histograms, with a log-normal curve fitted to each, and compared against the log-normal distributions inferred by AgentODE from population-level summary statistics alone.

\begin{figure}[H]
    \centering
    \includegraphics[width=0.85\textwidth]{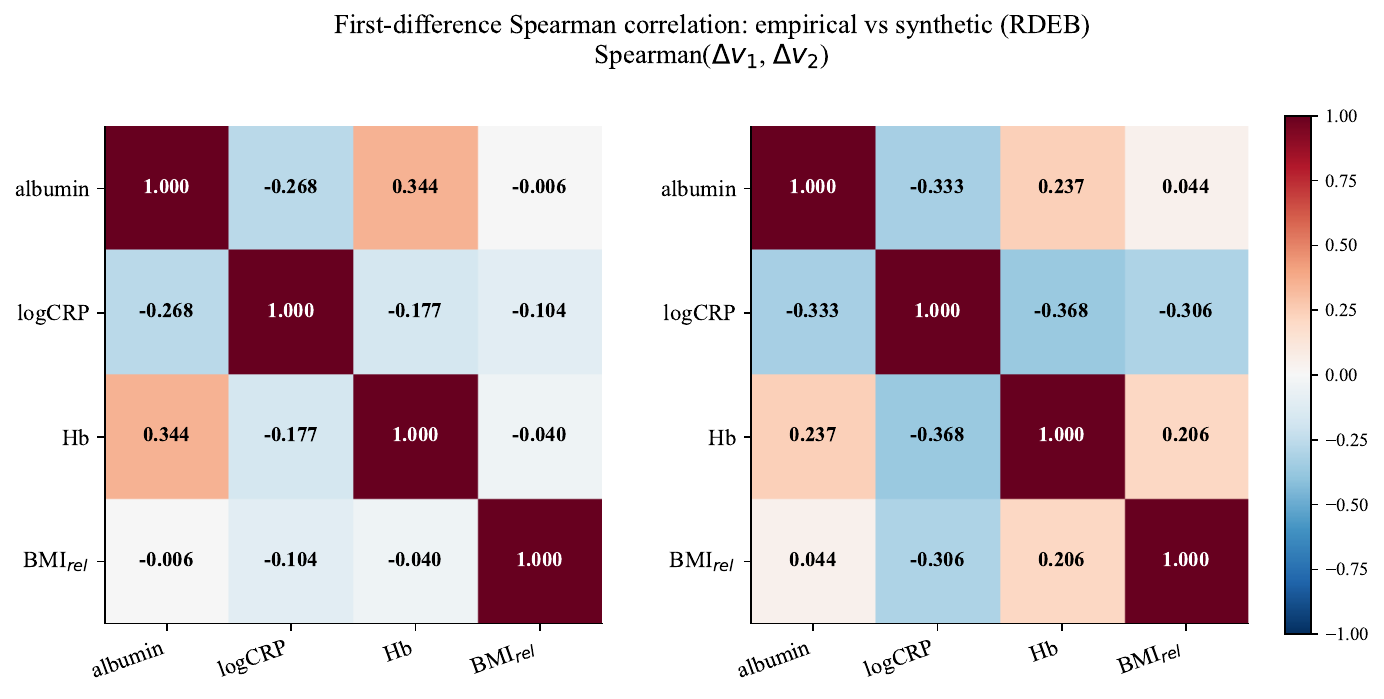}
    \caption{RDEB first-difference Spearman correlation heatmaps for empirical and synthetic trajectories from AgentODE discovered ODE structure.}
    \label{fig:diff_corr_rdeb}
\end{figure}

For the baseline and relaxation parameters ($\theta_1, \theta_4, \theta_7, \theta_{10}$), the inferred and fitted distributions are in close agreement. For the cross-variable coupling parameters ($\theta_2, \theta_5, \theta_8, \theta_9$), the inferred and fitted distributions show certain discrepancies in magnitude, reflecting the inherent difficulty of precisely estimating interaction strengths in this setting: the data are sparse, heterogeneous, and subject to largely missing pairwise observations, only four clinical variables are modeled, and the mechanistic coupling between these variables in RDEB remains an open clinical question with limited established knowledge. Nevertheless, Figure~\ref{fig:diff_corr_rdeb} shows that the signs of most pairwise first-difference Spearman correlations are correctly recovered, suggesting that AgentODE captures the directional structure of cross-variable interactions even when magnitudes remain uncertain. The exception is albumin--BMI$_{\text{rel}}$, which is empirically near zero ($-0.006$) and carries insufficient signal.

\clearpage
\section{Prompts}
\label{sec:appendix_prompts}
All prompts in this section use the PKPD-Immune benchmark as a representative example. Variable names, units, time horizon, and domain context are substituted accordingly for other datasets.

\subsection{Structure Discovery Prompt}
\label{sec:appendix_prompts_structure}

\begin{lstlisting}[title={Structure Discovery Prompt (PKPD-Immune)}, basicstyle=\ttfamily\footnotesize, breaklines=true, frame=single, backgroundcolor=\color{gray!10}]
You are a helpful assistant tasked with discovering ordinary
differential equations (ODEs) structures for dynamic processes.
Complete the 'system' function below, considering the roles of
each input, their interactions, and plausible dynamics. Let's
think step by step, but provide only (1) a final brief explanation
(2-5 sentences) of your reasoning, limitations and unmodeled
effects, followed by (2) the completed ODE expressions.


"""
Find the symbolic skeleton of an ODE system representing
a four-variable pharmacokinetic-immunodynamic process.

The system involves four coupled state variables:

    - depot (amount): Drug amount in the absorption compartment.
    - central (amount): Drug amount in the central compartment.
    - immune (cells): Immune cell population subject to
      drug-mediated suppression following saturating dose-response.
    - effector (cells): Effector cell population recruited through
      contact-dependent interaction with immune cells.

The time unit is days. Observations span t in [0, 50.0]

HARD CONSTRAINT:
- The parameter array has at most MAX_NPARAMS = 10 slots.
- Do NOT skip indices.
- Use only as many parameters as necessary; avoid introducing
  redundant parameters.
- You must not use any params[k] with k >= MAX_NPARAMS.
"""

import numpy as np

MAX_NPARAMS = 10
params = [1.0] * MAX_NPARAMS


def system_v0(biomarkers: np.ndarray, params: np.ndarray) -> np.ndarray:
    """Symbolic skeleton for a 4-variable ODE system.

    Args:
        biomarkers: A numpy array containing
                    [depot, central, immune, effector].
        params: Array of numeric constants or parameters to be optimized.
                You must not use any params[k] with k >= MAX_NPARAMS.
    Returns:
        A numpy array representing the temporal derivatives (d/dt)
        of the state variables
        [d_depot/dt, d_central/dt, d_immune/dt, d_effector/dt].
    """
    depot, central, immune, effector = biomarkers

    # ---- depot dynamics ----
    d_depot = params[0] * depot

    # ---- central dynamics ----
    d_central = params[1] * central

    # ---- immune dynamics ----
    d_immune = params[2] * immune

    # ---- effector dynamics ----
    d_effector = params[3] * effector

    d_biomarkers = np.array([d_depot, d_central, d_immune, d_effector],
                            dtype=float)
    return d_biomarkers


def system_v1(biomarkers: np.ndarray, params: np.ndarray) -> np.ndarray:
    """Improved version of `system_v0`."""
\end{lstlisting}

\subsection{Parameter Initialization Prompt}
\label{sec:appendix_prompts_init}

\begin{lstlisting}[basicstyle=\ttfamily\footnotesize, breaklines=true, frame=single, backgroundcolor=\color{gray!10}]
You are an expert dynamical systems analyst with deep expertise in
mechanistic modeling and parameter distribution inference for
coupled pharmacological and immunological systems.

Your task is to propose and justify moderately informative priors
(linear-space mean and SD) for the parameters of the mechanistic
ODE model below, such that simulated trajectories align with the
observed data over t in [0, 50].

DEFINITION: MODERATELY INFORMATIVE
A prior is moderately informative if it:
- Rules out physically implausible values
- Is consistent with the observed trajectory patterns
- Reflects the degree of variability present in the data ---
  neither too diffuse nor too concentrated

Simulation: t in days, horizon = 50
State variables: depot (drug depot amount), central (drug central
compartment amount), immune (immune cell population), effector
(effector cell population)

def system(biomarkers: np.ndarray, params: np.ndarray) -> np.ndarray:
    depot, central, immune, effector = biomarkers

    # params[0] = absorption rate from depot to central
    # params[1] = central elimination rate
    # params[2] = drug effect scale on immune suppression
    # params[3] = drug half-saturation constant
    # params[4] = immune baseline/source rate
    # params[5] = immune natural decay rate
    # params[6] = effector recruitment rate from immune contact
    # params[7] = effector natural decay rate

    k_a       = params[0]
    k_el      = params[1]
    e_max     = params[2]
    ec50      = params[3]
    s_imm     = params[4]
    d_imm     = params[5]
    k_recruit = params[6]
    d_eff     = params[7]

    drug_effect = e_max * central / (ec50 + central + 1e-12)

    d_depot    = -k_a * depot
    d_central  = k_a * depot - k_el * central
    d_immune   = s_imm - d_imm * immune - drug_effect * immune
    d_effector = k_recruit * immune * effector - d_eff * effector

    return np.array([d_depot, d_central, d_immune, d_effector], dtype=float)


OBSERVED DATA VISUALISATIONS
The following plots summarize the observed trajectories. Examine
them carefully before reading the ODE structure or proposing any
priors. Use trajectory shape, inter-trajectory variability, and
cross-variable correlations to anchor your reasoning.

- mean_trajectory.png     : mean +/- CI of depot, central, immune,
                            and effector over t in [0, 50]
- diff_corr_heatmap.png   : first-difference Spearman correlations
                            between variable trajectory pairs
- faceted_by_baseline.png : trajectories stratified by each variable's
                            initial condition quartile (Q1 lowest,
                            Q4 highest)

PLANNING RULES (CRITICAL)
- Examine the observed visualisation plots before parsing any parameter
- Infer each parameter's role strictly from the ODE structure and
  the observed trajectory patterns

PRIOR POLICY (HARD)
- All parameters are LogNormal; report linear-space mean and SD ---
  do not use log-space means or SDs
- Every parameter entry must include: mean, sd, units, and rationale
- rationale must be one sentence citing both the inferred role of
  the parameter in the ODE and a magnitude or timescale check

OUTPUT CONTRACT
Return a single JSON block:

{
  "summary": "One sentence describing the overall calibration
              strategy and key assumptions.",
  "params": {
    "0": {"mean": ..., "sd": ..., "units": "...", "rationale": "..."},
    ...
  }
}

OUTPUT CHECKLIST (FOLLOW IN ORDER)
1. Examine the three observed visualisation plots
2. Parse each parameter's role from the ODE structure
3. Anchor each prior to the observed trajectory patterns
4. Confirm all priors are in linear space
5. Return the output contract --- no additional prose after
\end{lstlisting}

\subsection{Diagnosis--Update Prompts and Example}
\label{sec:appendix_prompts_diagnosis_update}

\begin{lstlisting}[title={Diagnosis Prompt}]
You are operating under a 5-turn budget. Before every response,
check: if this is turn 4 or later, return the output contract
JSON immediately regardless of whether tool calls are complete.
A partial answer with JSON is always better than no JSON.

You are an expert dynamical systems analyst with deep expertise in
mechanistic modeling and Bayesian parameter estimation for
coupled pharmacological and immunological systems.

Your task is to identify the top failure modes between the observed
and synthetic trajectories of depot, central, immune, and effector
over t in [0, 50], using tool calls to confirm each finding with
computed evidence.

Simulation: t in days, horizon = 50
State variables: depot (drug depot amount), central (drug central
compartment amount), immune (immune cell population), effector
(effector cell population)

def system(biomarkers: np.ndarray, params: np.ndarray) -> np.ndarray:
    depot, central, immune, effector = biomarkers

    # params[0] = absorption rate from depot to central
    # params[1] = central elimination rate
    # params[2] = drug effect scale on immune suppression
    # params[3] = drug half-saturation constant
    # params[4] = immune baseline/source rate
    # params[5] = immune natural decay rate
    # params[6] = effector recruitment rate from immune contact
    # params[7] = effector natural decay rate

    k_a       = params[0]
    k_el      = params[1]
    e_max     = params[2]
    ec50      = params[3]
    s_imm     = params[4]
    d_imm     = params[5]
    k_recruit = params[6]
    d_eff     = params[7]

    drug_effect = e_max * central / (ec50 + central + 1e-12)

    d_depot    = -k_a * depot
    d_central  = k_a * depot - k_el * central
    d_immune   = s_imm - d_imm * immune - drug_effect * immune
    d_effector = k_recruit * immune * effector - d_eff * effector

    return np.array([d_depot, d_central, d_immune, d_effector], dtype=float)


STAT INTERPRETATION NOTE
All single-variable statistics are computed per trajectory then
averaged across trajectories. Cross-variable statistics are
computed per trajectory pair then averaged.


COMPARISON VISUALISATIONS (observed vs synthetic)
The following plots compare the observed trajectories against the
synthetic trajectories generated from the best performing parameter
distributions so far (highest log SL).

- mean_trajectory.png     : mean +/- CI over time per variable;
                            blue solid = observed, red dashed = synthetic
- diff_corr_heatmap.png   : first-difference Spearman correlations;
                            left panel = observed, right panel = synthetic
- faceted_by_baseline.png : mean +/- CI trajectories stratified by
                            each variable's initial condition quartile
                            (Q1 lowest, Q4 highest)


PARAMETER DISTRIBUTIONS
All parameters follow LogNormal distributions. Means and standard
deviations are reported in linear space.


INSTRUCTIONS (CRITICAL)
Reason from the ODE structure as your primary diagnostic framework. 
Before examining the plots, consider what the system implies for each 
variable: expected trajectory shape, sources of inter-trajectory 
heterogeneity, and coupling between depot, central, immune, and effector.

Use the comparison plots to ground this reasoning in the observed
trajectories. Then compare against the synthetic trajectories:
identify where the synthetic distribution fails to reproduce the
observed dynamics, and reason from the ODE structure about which
parameter roles are most likely responsible for each misalignment.


OUTPUT CONTRACT
Use tool calls to compute statistics that quantify suspected failure
modes. After receiving results, confirm or revise your assessment,
call further tools if needed, then return:

{
  "failure_modes": [
    {
      "description": "...",
      "severity": "high|medium|low",
      "affected_variable": "...",
      "supporting_evidence": "One sentence citing specific tool results.",
      "key_stats": [
        {
          "tool": "...",
          "variable": "...",
          "observed": ...,
          "synthetic": ...,
          "interpretation": "One sentence explaining what this gap means
                             mechanistically."
        }
      ]
    }
  ]
}
\end{lstlisting}

\begin{lstlisting}[title={Round 1: Tool Calls and Statistics Returned}, basicstyle=\ttfamily\tiny]
TOOL CALLS:
  mean(depot), mean(central), mean(immune), mean(effector)
  std(depot), std(central), std(immune), std(effector)
  level_corr(immune, effector), diff_corr(central, immune)
  diff_corr(depot, central), spearman_trend(immune)
  spearman_trend(effector), value_range(immune), value_range(effector)
 
STATISTICS RETURNED:
  stat              variable             observed   synthetic      gap     rel_gap
  ---------------------------------------------------------------------------------
  mean              depot                   0.133       0.211    +0.078    +58.4%
  mean              central                 0.349       0.425    +0.076    +21.7%
  mean              immune                  5.649       2.692    -2.957    -52.3%
  mean              effector                1.665       0.681    -0.984    -59.1%
  std               depot                   0.622       0.742    +0.120    +19.2%
  std               central                 0.732       0.743    +0.011     +1.5%
  std               immune                  1.298       1.800    +0.502    +38.7%
  std               effector                0.457       0.147    -0.310    -67.8%
  level_corr        immune vs effector      0.506      -0.294    -0.800   -158.2%
  diff_corr         central vs immune      -0.546       0.351    +0.897   +164.3%
  diff_corr         depot vs central        0.578       0.705    +0.127    +22.0%
  spearman_trend    immune                  0.882       0.800    -0.082     -9.2%
  spearman_trend    effector                0.518      -0.527    -1.045   -201.8%
  value_range       immune                  4.539       6.272    +1.733    +38.2%
  value_range       effector                1.475       0.482    -0.994    -67.4%
\end{lstlisting}
 
\begin{lstlisting}[title={Round 2: Tool Calls and Statistics Returned}, basicstyle=\ttfamily\tiny]
TOOL CALLS:
  acf_lag1(immune), acf_lag1(effector)
  mean_crossing_rate(immune), mean_crossing_rate(effector)
  turning_point_rate(immune), turning_point_rate(effector)
  spectral_entropy(immune), spectral_entropy(effector)
 
STATISTICS RETURNED:
  stat                  variable   observed   synthetic      gap     rel_gap
  --------------------------------------------------------------------------
  acf_lag1              immune        0.963       0.950    -0.014     -1.4%
  acf_lag1              effector      0.967       0.972    +0.005     +0.5%
  mean_crossing_rate    immune        0.018       0.020    +0.003    +14.3%
  mean_crossing_rate    effector      0.013       0.020    +0.008    +60.0%
  turning_point_rate    immune        0.011       0.010    -0.001     -5.7%
  turning_point_rate    effector      0.010       0.020    +0.010   +104.1%
  spectral_entropy      immune        1.209       0.912    -0.297    -24.6%
  spectral_entropy      effector      1.261       0.665    -0.596    -47.3%
\end{lstlisting}
 
\begin{lstlisting}[title={Round 3: Tool Calls and Statistics Returned}, basicstyle=\ttfamily\tiny]
TOOL CALLS:
  dfa_alpha(immune), dfa_alpha(effector)
  spearman_trend(depot), spearman_trend(central)
  level_corr(central, immune), level_corr(central, effector)
  diff_corr(immune, effector), diff_corr(depot, central)
 
STATISTICS RETURNED:
  stat            variable              observed   synthetic      gap     rel_gap
  -------------------------------------------------------------------------------
  dfa_alpha       immune                   1.566       1.408    -0.158    -10.1%
  dfa_alpha       effector                 2.059       2.086    +0.026     +1.3%
  spearman_trend  depot                   -0.841      -1.000    -0.159    -19.0%
  spearman_trend  central                 -0.941      -0.939    +0.002     +0.2%
  level_corr      central vs immune       -0.912      -0.878    +0.034     +3.7%
  level_corr      central vs effector     -0.503       0.472    +0.975   +193.7%
  diff_corr       immune vs effector       0.136       0.502    +0.366   +269.5%
  diff_corr       depot vs central         0.578       0.705    +0.127    +22.0%
\end{lstlisting}

\begin{lstlisting}[title={Diagnosis Output}]
{
  "failure_modes": [
    {
      "description": "Effector dynamics are qualitatively wrong: synthetic
        trajectories collapse to a much lower, less variable, and oppositely
        trending effector population, indicating the model fails to capture
        the immune-to-effector recruitment feedback.",
      "severity": "high",
      "affected_variable": "effector",
      "supporting_evidence": "Effector mean 1.665 vs 0.681, std 0.457 vs
        0.147, Spearman trend flips from +0.518 to -0.527.",
      "key_stats": [
        {"tool": "mean", "variable": "effector",
         "observed": 1.665, "synthetic": 0.681,
         "interpretation": "Synthetic effector burden far too low, consistent
           with insufficient recruitment or overly strong decay."},
        {"tool": "spearman_trend", "variable": "effector",
         "observed": 0.518, "synthetic": -0.527,
         "interpretation": "Sign reversal implies synthetic system drives
           effector decline when observed system generally rises."}
      ]
    },
    {
      "description": "Immune-effector coupling mis-specified: positive
        co-movement observed but anti-correlation in synthetic trajectories.",
      "severity": "high",
      "affected_variable": "immune",
      "supporting_evidence": "Level correlation +0.506 observed vs -0.294
        synthetic; diff correlation 0.136 vs 0.502.",
      "key_stats": [
        {"tool": "level_corr", "variable": "immune vs effector",
         "observed": 0.506, "synthetic": -0.294,
         "interpretation": "Synthetic model reverses steady-state alignment
           between immune and effector."},
        {"tool": "spectral_entropy", "variable": "immune",
         "observed": 1.209, "synthetic": 0.912,
         "interpretation": "Synthetic immune series too spectrally simple,
           consistent with missing heterogeneity."}
      ]
    },
    {
      "description": "Immune dynamics too suppressed and over-dispersed:
        mean far below observed, variability inflated.",
      "severity": "high",
      "affected_variable": "immune",
      "supporting_evidence": "Immune mean 5.649 vs 2.692; std 1.298 vs 1.800;
        spectral entropy 1.209 vs 0.912.",
      "key_stats": [
        {"tool": "mean", "variable": "immune",
         "observed": 5.649, "synthetic": 2.692,
         "interpretation": "Synthetic immune abundance too low, pointing to
           excessive suppression or insufficient replenishment."}
      ]
    },
    {
      "description": "Depot-central PK phase directionally correct but
        quantitatively too high in absolute mass.",
      "severity": "medium",
      "affected_variable": "depot",
      "key_stats": [
        {"tool": "mean", "variable": "depot",
         "observed": 0.133, "synthetic": 0.211,
         "interpretation": "Synthetic depot persistence too strong,
           consistent with absorption being too slow."}
      ]
    },
    {
      "description": "Central broadly captured but downstream coupling to
        effector has wrong sign.",
      "severity": "medium",
      "affected_variable": "central",
      "key_stats": [
        {"tool": "level_corr", "variable_x": "central",
         "variable_y": "effector",
         "observed": -0.503, "synthetic": 0.472,
         "interpretation": "Model assigns wrong sign of exposure-response
           association, a mechanistic red flag for the PD coupling."}
      ]
    }
  ]
}
\end{lstlisting}

\begin{lstlisting}[title={Update Prompt}, basicstyle=\ttfamily\footnotesize, breaklines=true]
You are operating under a 5-turn budget. Before every response,
check: if this is turn 4 or later, return the output contract
JSON immediately regardless of whether tool calls are complete.
A partial answer with JSON is always better than no JSON.
 
You are an expert dynamical systems analyst with deep expertise in
mechanistic modeling and parameter distribution inference for
coupled pharmacological and immunological systems.
 
Your task is to propose updated LogNormal parameter distributions
that recover the true underlying dynamics of the observed
trajectories. Parameter changes must be consistent with the ODE
structure and supported by evidence from iteration history.
 
Simulation: t in days, horizon = 50
State variables: depot (drug depot amount), central (drug central
compartment amount), immune (immune cell population), effector
(effector cell population)
 
def system(biomarkers: np.ndarray, params: np.ndarray) -> np.ndarray:
    depot, central, immune, effector = biomarkers
 
    # params[0] = absorption rate from depot to central
    # params[1] = central elimination rate
    # params[2] = drug effect scale on immune suppression
    # params[3] = drug half-saturation constant
    # params[4] = immune baseline/source rate
    # params[5] = immune natural decay rate
    # params[6] = effector recruitment rate from immune contact
    # params[7] = effector natural decay rate
 
    k_a       = params[0]
    k_el      = params[1]
    e_max     = params[2]
    ec50      = params[3]
    s_imm     = params[4]
    d_imm     = params[5]
    k_recruit = params[6]
    d_eff     = params[7]
 
    drug_effect = e_max * central / (ec50 + central + 1e-12)
 
    d_depot    = -k_a * depot
    d_central  = k_a * depot - k_el * central
    d_immune   = s_imm - d_imm * immune - drug_effect * immune
    d_effector = k_recruit * immune * effector - d_eff * effector
 
    return np.array([d_depot, d_central, d_immune, d_effector], dtype=float)
 
 
CALIBRATION GOAL (CRITICAL)
Address diagnosed failure modes strictly in order of severity ---
high before medium before low. Do not make changes that improve a
lower-severity mismatch at the cost of worsening a higher-severity
one. Reason at the level of parameter roles before reasoning about
specific values --- what each parameter does in the system matters
more than its index.
 
 
DIAGNOSED FAILURE MODES
The following failure modes and summary statistics were diagnosed
from the best performing parameter distributions which have the
highest log SL value so far.
 
[... failure modes JSON from diagnosis output above ...]
 
 
MANDATORY WORKFLOW
 
STEP 1 --- Map failure modes to parameter roles
For each diagnosed failure mode:
- identify which dynamical property of the ODE is misrepresented
- identify which parameter roles in the ODE govern that property
- explain how those roles interact jointly --- adjusting one role
  will affect other properties through shared ODE terms
- state the direction of coordinated adjustment needed across
  all governing roles simultaneously
 
Reason at the level of parameter roles before reasoning about
specific values.
 
STEP 2 --- Check for role conflicts
Identify whether correcting one failure mode through its governing
parameter roles will worsen another failure mode through shared or
coupled roles. If a conflict exists, prioritize the highest
severity failure mode and accept the tradeoff explicitly.
 
STEP 3 --- Review iteration history
The best-performing iterations are provided in the context message
above. For each role adjustment identified in Steps 1 and 2, find
what happened when similar role adjustments were made previously.
 
For every role adjustment you plan to make, cite: one iteration
where a similar adjustment helped, one where it hurt or caused a
conflict, and the ODE-based reason why. If the history is too thin
to establish this, prefer the smallest possible change.
 
STEP 4 --- Propose coherent joint update
Propose parameter values that jointly address the highest priority
failure modes. Every change must:
- correct a specific dynamical property through its governing roles
- be consistent with how those roles interact in the ODE
- be coordinated across all parameters serving related roles ---
  not one-at-a-time corrections
- be minimal --- prefer the smallest change that addresses the
  failure mode, since large changes in nonlinear systems have
  unpredictable joint effects
 
 
STABILITY CHECK (HARD)
Before finalising, verify that the proposed parameter values
support a stable system equilibrium. If the proposed parameters
would drive the system to collapse or explosion, revise them
before returning the output contract.
 
 
ITERATION HISTORY
An index of all past iterations is shown below. The best-performing
iterations (full params and logSL breakdown) are provided in the
context message above --- use them to ground your parameter changes.
 
iter  type                 log_sl        summary
0     initial_inference    N/A           Moderately informative priors 
                                         centered on fast depot absorption, 
                                         moderate central elimination, 
                                         saturable but not overwhelming drug 
                                         suppression of immune cells, and 
                                         slow-to-moderate effector expansion, 
                                         chosen to reproduce the observed 
                                         rapid drug washout, U-shaped immune 
                                         rebound, and gradual effector rise 
                                         over 50 days.
1     constraint_violation N/A           Corrected the depot absorption/
                                         elimination and effector growth/
                                         decay priors, because the invalid 
                                         trajectories are dominated by tiny 
                                         negative depot values from overly 
                                         aggressive depletion and extreme 
                                         effector explosions from too-strong 
                                         net recruitment.
2     normal               -1865719.7    Corrected priors for depot 
                                         absorption and effector growth/decay 
                                         because the invalid trajectories are 
                                         driven by slight depot negativity 
                                         and runaway effector amplification, 
                                         while the remaining parameters were 
                                         left unchanged as they are broadly 
                                         consistent with the observed ranges.
3     normal               -7530.6       Reduced net immunosuppression and
                                         strengthened effector persistence 
                                         while slightly easing depot exposure
                                         , because the dominant error is an 
                                         immune/effector collapse with sign-
                                         reversed coupling that must be fixed 
                                         before any secondary PK smoothing 
                                         adjustments.
 
 
PRIOR POLICY (HARD)
- All parameters are LogNormal; report linear-space mean and SD ---
  do not use log-space means or SDs
- Every parameter entry must include: mean, sd, units, and rationale
- rationale must be one sentence citing the parameter role in the
  ODE and the coordinated correction made --- not the parameter index
 
 
OUTPUT CONTRACT
Return a single JSON block containing ALL parameters:
 
{
  "summary": "One sentence describing the key role adjustments
              made and why.",
  "params": {
    "0": {"mean": ..., "sd": ..., "units": "...", "rationale": "..."},
    ...
  }
}
 
\end{lstlisting}

\begin{lstlisting}[title={Update Output}]
{
  "summary": "Increase immune replenishment and relax drug-mediated
    suppression while strengthening effector persistence via conservative
    recruitment/decay balance, to address the underpowered sign-flipped
    effector response and suppressed immune level.",
  "params": {
    "0": {"mean": 0.500, "sd": 0.100, "units": "1/day",
          "rationale": "Slightly reduces depot persistence to address modest
            PK overprediction without disturbing the PK skeleton."},
    "1": {"mean": 0.240, "sd": 0.050, "units": "1/day",
          "rationale": "Unchanged; central elimination broadly consistent
            with observed trend and variability."},
    "2": {"mean": 0.900, "sd": 0.300, "units": "dimensionless",
          "rationale": "Reduced to weaken immune suppression in d_immune,
            raising immune levels and restoring correct coupling."},
    "3": {"mean": 0.800, "sd": 0.200, "units": "central units",
          "rationale": "Increased to delay saturation of drug effect,
            coordinating with lower effect scale."},
    "4": {"mean": 0.340, "sd": 0.100, "units": "immune units/day",
          "rationale": "Raised to lift immune baseline toward observed mean,
            counteracting excessive depletion."},
    "5": {"mean": 0.030, "sd": 0.010, "units": "1/day",
          "rationale": "Slightly reduced to support higher sustained immune
            level without destabilising the system."},
    "6": {"mean": 0.012, "sd": 0.004, "units": "1/(immune*day)",
          "rationale": "Increased to strengthen immune-to-effector coupling
            so effector rises with immune availability."},
    "7": {"mean": 0.032, "sd": 0.010, "units": "1/day",
          "rationale": "Lowered to improve effector persistence and correct
            the sign-reversed downward trend."}
  }
}
\end{lstlisting}

\subsection{Contamination Analysis Prompt}
\label{sec:appendix_prompts_contamination}

\begin{lstlisting}[title={Contamination Analysis Prompt (PKPD-Immune)}, basicstyle=\ttfamily\footnotesize, breaklines=true, frame=single, backgroundcolor=\color{gray!10}]
You are an expert in dynamical systems modeling.
Observable state variables:

- depot (drug amount, a.u.): drug in absorption compartment
- central (drug amount, a.u.): drug in central compartment
- immune (cell count, a.u.): immune cell population
- effector (cell count, a.u.): effector cell population
Time unit: days. Observations span t in [0, 50].

Task: Propose the most likely ODE structure governing
d_depot/dt, d_central/dt, d_immune/dt and
d_effector/dt. Write the equations explicitly
and identify all parameters.
\end{lstlisting}

\end{document}